\documentclass{article}

\usepackage{arxiv}

\usepackage[utf8]{inputenc} 
\usepackage[T1]{fontenc}    
\usepackage{hyperref}       
\usepackage{url}            
\usepackage{booktabs}       
\usepackage{amsfonts}       
\usepackage{nicefrac}       
\usepackage{microtype}      
\usepackage{lipsum}
\usepackage{subfigure}
\usepackage{graphicx}
\usepackage[ruled,vlined,noresetcount]{algorithm2e}
\usepackage{amsmath}

\title{Long-Bone Fracture Detection using Artificial Neural Networks based on Contour Features of X-ray Images}

\author{
  Alice Yi Yang \\ 
  School of Electrical and Information Engineering \\
  University of the Witwatersrand \\
  Johannesburg, South Africa, 2000 \\
  \texttt{yangalice8@gmail.com} \\
   \And
 Ling Cheng \\
  School of Electrical and Information Engineering\\
  University of the Witwatersrand\\
  Johannesburg, South Africa, 2000 \\
  \texttt{ling.cheng@wits.ac.za} \\
}

\begin{document}
\maketitle

\begin{abstract}
The following paper proposes two contour-based fracture detection schemes. The schemes are intended to assist medical professionals within the medical field in fracture detection of the human long-bone from X-ray images. The development of the contour-based fracture is based on the line-based fracture detection schemes proposed in \cite{yang_long-bone_2019}. Existing Computer Aided Diagnosis (CAD) systems commonly employs Convolutional Neural Networks (CNN) for the classification of fractured X-ray images. Although the existing CAD systems obtain a high accuracy, the cost for a high accuracy is the amount of training data required. The purpose of the proposed schemes is to obtain a high classification accuracy with a reduced number of training data through the use of detected contours in X-ray images. There are two contour-based fracture detection schemes. The first is the Standard Contour Histogram Feature-Based (CHFB) and the second is the improved CHFB scheme. The difference between the two schemes is the removal of the surrounding detected flesh contours from the leg region in the improved CHFB scheme. The flesh contours are automatically classified as non-fractures. The contours are further refined to give a precise representation of the image edge objects. A total of 19 features are extracted from each refined contour. 8 out of the 19 features are based on the number of occurrences for particular detected gradients in the contour. Moreover, the occurrence of the $0^\circ$ gradient in the contours are employed for the separation of the knee, leg and foot region. The features are a summary representation of the contour, in which it is used as inputs into the Artificial Neural Network (ANN). The system is evaluated with two evaluations. The first is an overall system performance basis, that evaluates the system with contours that have context about the image it is extracted from. The second evaluation only evaluates the ANN of the system with randomly selected contours without context about the image it is extracted from. Both Standard CHFB and improved CHFB schemes are evaluated with the same experimental set-ups. The average system accuracy for the Standard CHFB scheme is 80.7\%, whilst the improved CHFB scheme has an average accuracy of 82.98\%. Additionally, the hierarchical clustering technique is adopted to highlight the fractured region within the X-ray image, using extracted $0^\circ$ gradients from fractured contours.
\end{abstract}


\section{Introduction}
Within the medical field, an emphasis is placed on making an accurate detection for both medical professionals and Computer Aided Diagnosis systems (CAD). In \cite{DOI2007198}, the author looks at the history of CAD systems and current status of CAD systems. The paper looks at CAD systems that detect the conditional properties for specific medical conditions. These are: lung nodules, interstitial opacities, cardiomegaly, vertebral fractures, and interval changes in chest radiographs. The classification of the CAD systems are reported as being a "second opinion" before a final decision is made by physicians. For common CAD systems that employs artificial neural networks, the cost for an accurate image-based diagnosis is amount of data required to train the neural network. Kim, D.H Et al. \cite{kim2018artificial} proposes fracture detection model to investigate the extent of which learning can be transferred in CNN networks. The fracture detection focused on detecting wrist fractures from X-ray images. The CNN is constructed with five layers: one input and output layer, two hidden layers and connected layer. The authors trained the model with a total of 11,112 images. The training includes images that were labelled, both fractured and non-fractured. A total of 100 images were used for the testing of the model. The images consisted of 50 fractured and 50 no-fractured images. The results obtained by the authors indicated that the sensitivity and specificity resulted in values of 0.9 and 0.88, respectively. Additionally the AUC value obtained was 0.954, indicating that the model proposed has a high sensitivity detection for fractures. Brahim A. Et al. \cite{brahim_decision_2019} presents a CAD system for early knee Osteo Arthritis (OA) detection. The system utilises knee X-ray imaging and machine learning algorithms for the detection. The X-ray images are pre-processed using the Fourier filter in the Fourier domain. The selection from the extracted features are performed using Independent Component Analysis (ICA), which is a dimensionality reduction technique. These selected features are given to the Na\"ive Bayes and random forest classifiers for the classification of OA detection. The authors achieved an 82.98\% accuracy using 1024 knee X-ray images. Feature extraction is essential for classification of particular conditions. In \cite{wang_lung_2017}, a CAD system using deep feature fusion is proposed for the detection of lung nodules. Features are a key step to CAD systems, the authors fuse a variety of features obtained from varying CAD systems which employs the classic Convolutional Neural Networks (CNN). Principal Component Analysis (PCA) is applied to features for dimensionality reduction, hence only critical features are considered. The authors obtained improved results using the deep feature fusion for detection of lung nodules compared to the classic CNN.

In this paper, two Contour Histogram Feature-Based (CHFB) fracture detection scheme is proposed: Standard CHFB shceme and improved CHFB scheme. The schemes are based off of the line-based fracture detection schemes proposed in \cite{yang_long-bone_2019}. The difference between the two schemes is the automated detection of contours from the surround flesh in the improved CHFB scheme. The schemes adopts a feature based approach, in which the features are extracted from the detected contours found within the X-ray image. Thus, reducing the number of data images required for an accurate fracture detection. A total of 19 features are extracted from each contour. The extracted features are discussed in Section \ref{Section: Edge Feature Extraction}. The features are analysed using PCA to determine the dominant contour feature(s). The dominant feature(s) indicates the feature(s) that holds the most information to provide insight of the feature(s) which differentiates a fractured contour from a non-fractured contour. The application of PCA is detailed in Section \ref{Section: Contour Principle Component Analysis}. Before feature extraction, the detected contours are refined to eliminate duplicated information. The refinement of the contours is detailed in Section \ref{Section: Edge Extraction}. The additional information provided by the contour, allows for image segmentation. There are three distinct regions within an X-ray image, namely knee, leg and foot region. The image segmentation process separates the three regions, such that the focus is on the leg region. The image segmentation process is detailed in Section \ref{Section: Image Segmentation}. The ANN architecture for CHFB fracture detection consists of the four layers: one input and output layer, and two hidden layers. It has a similar ANN architecture to the ANN for both the Standard and ADPO schemes. However the difference is the number of nodes in each layer of the ANN. This is detailed in Section \ref{Section: CHFB Neural Network Architecture}. The results of the ANN are illustrated in Section \ref{Section: CHFB Neural Network Results}, where the experimental set-ups for the CHFB fracture detection scheme are discussed.

\section{Methodology}
\subsection{Contour Histogram Feature-Based Fracture Detection System Overview} \label{Section: Contour-Based System Overview}
\begin{figure}[ht!]
	\begin{center}
		\includegraphics[scale=0.8]{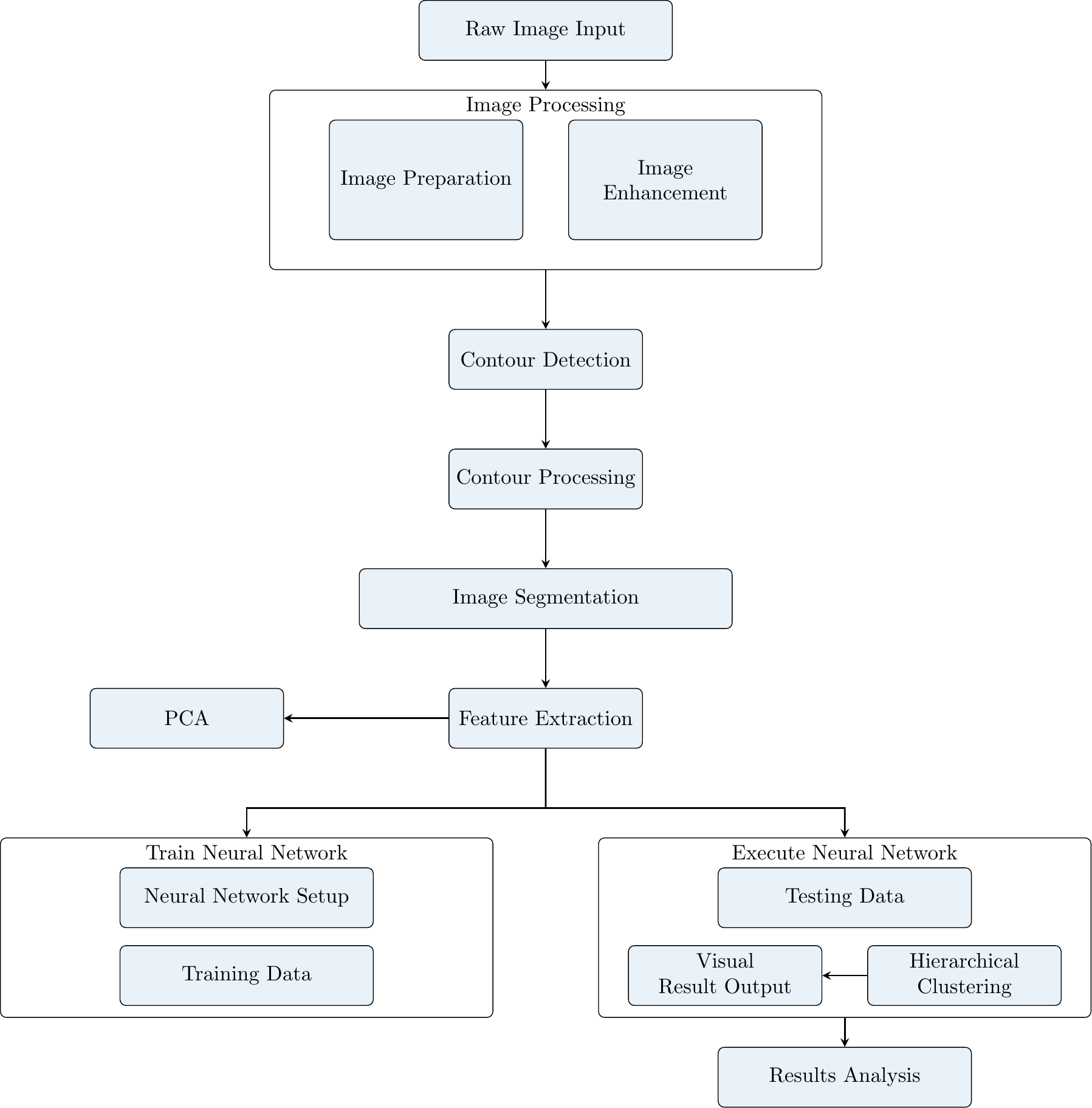}
	\end{center}
	\caption{Flowchart illustrating an overview of the system for the contour-based fracture detection scheme}
	\label{fig: contour-based system overview}
\end{figure}

The CHFB fracture detection scheme consists of similar components as both the Standard and ADPO schemes. The difference in schemes is the CHFB scheme consists of a contour detection component rather than a line detection, further contouring refinement and an image segmentation component is introduced. PCA is applied to the extracted features to determine the contributing dominant feature(s) for fractured and non-fractured contours. The training of the ANN consists of two components, namely the ANN architecture and the training data used to train the ANN. The execution of the ANN, both tests the accuracy of the ANN performance as well as obtains visual output results of the detected fractured contours. Additionally, the fractured region is identified using a hierarchical clustering technique on $0^\circ$ gradient points extracted from fractured contours. An overview of the CHFB fracture detection scheme is illustrated in Figure \ref{fig: contour-based system overview}.

\subsection{Image Enhancement}
The raw X-ray images are processed to ensure that the quality of the images are enhanced and constant before edge detection and contour extraction. The image enhancement process entails removal of surrounding white space, pixel equalisation. gamma correction, denoising and unsharp masking. The process ensures that the focus of the X-ray image is of the long-bone and all pixels are intensified to create a high contrast between the edges of the bones and all other pixels. The enhanced image is employed to generate a binary image (image containing black and white pixels only) with the bone edges highlighted. The edge detection is performed using the Canny edge detection technique. The tool box used for Canny edge detection is obtained from the OpenCV2.4 library \cite{Canny_Edge_Detection, green2002canny}.

\subsection{Contour Extraction} \label{Section: Contour Extraction}
\begin{figure}[ht!]
	\centering
	\subfigure[Original Enhanced X-ray Image]{\label{fig: Original Enhanced Image}\includegraphics[width=0.49\textwidth]{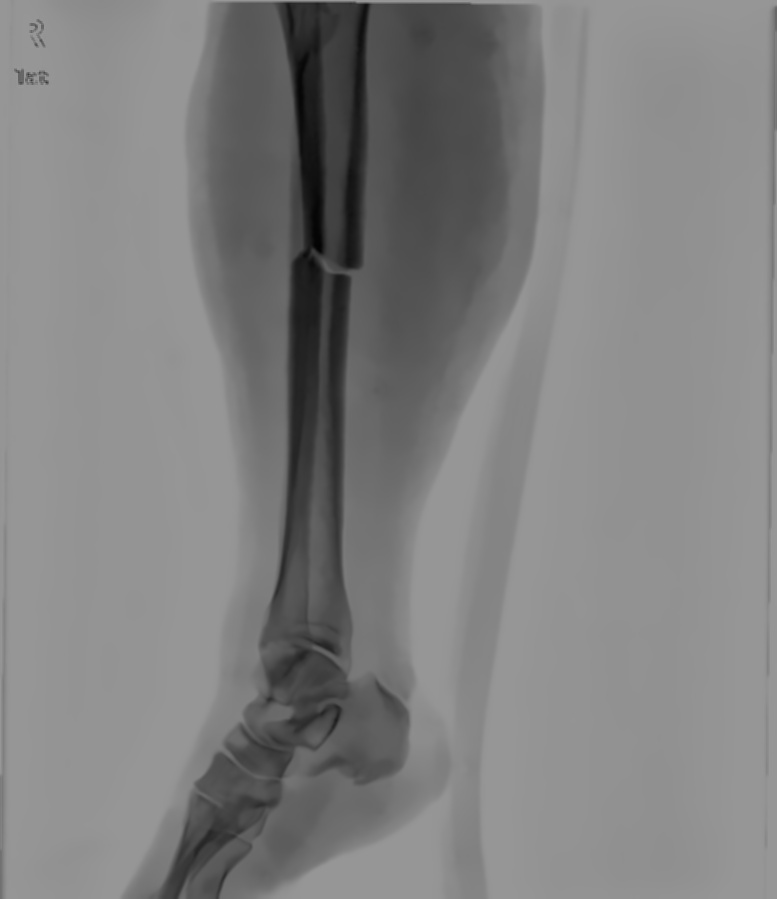}}
	\subfigure[Extracted Contours from X-ray Image]{\label{fig: extracted contours}\includegraphics[width=0.49\textwidth]{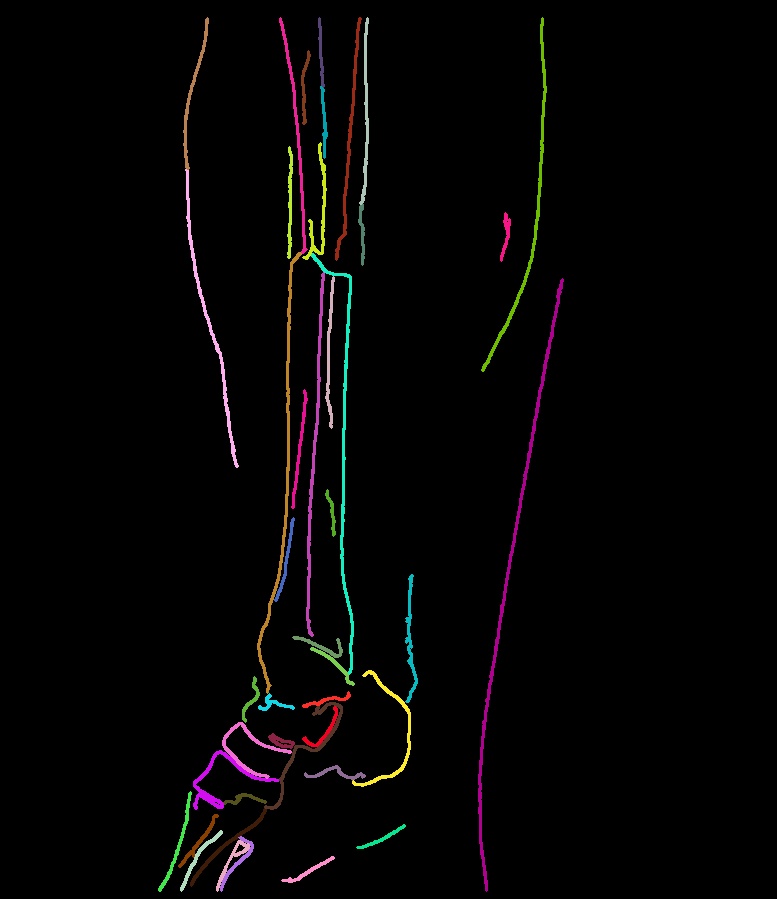}}
	\caption{Images illustrating the extracted contours from the enhanced X-ray image} 
	\label{fig: contour extraction}
\end{figure}

The contours are detected using the canny image generated from the Canny edge detection technique. The canny image consists of highlighted edges against a dark background. Contours defined as a sequence of points, $(x,y)$ surrounding image edge objects \cite{FindContours}. Therefore each detected contour varies in size depending on the size of the image edge object. The size of the contour is determined by the number of points that construct the contour. The contour provides more information about the image edge objects compared to the detected lines. This is because contours are composed of more than two points to represent image edge objects found within the X-ray image. However, since the contour surrounds the edge object there are repeated points that introduces redundancy. Consequently, the contours are further refined to eliminate the redundancy as the redundant points distorts the representation of the image edge objects. Figure \ref{fig: contour extraction} illustrates the extracted contours from the X-ray image. 

\subsection{Contour Refinement Processing} \label{Section: Edge Extraction}
\begin{figure}[h!]
	\begin{center}
		\includegraphics[scale=0.67]{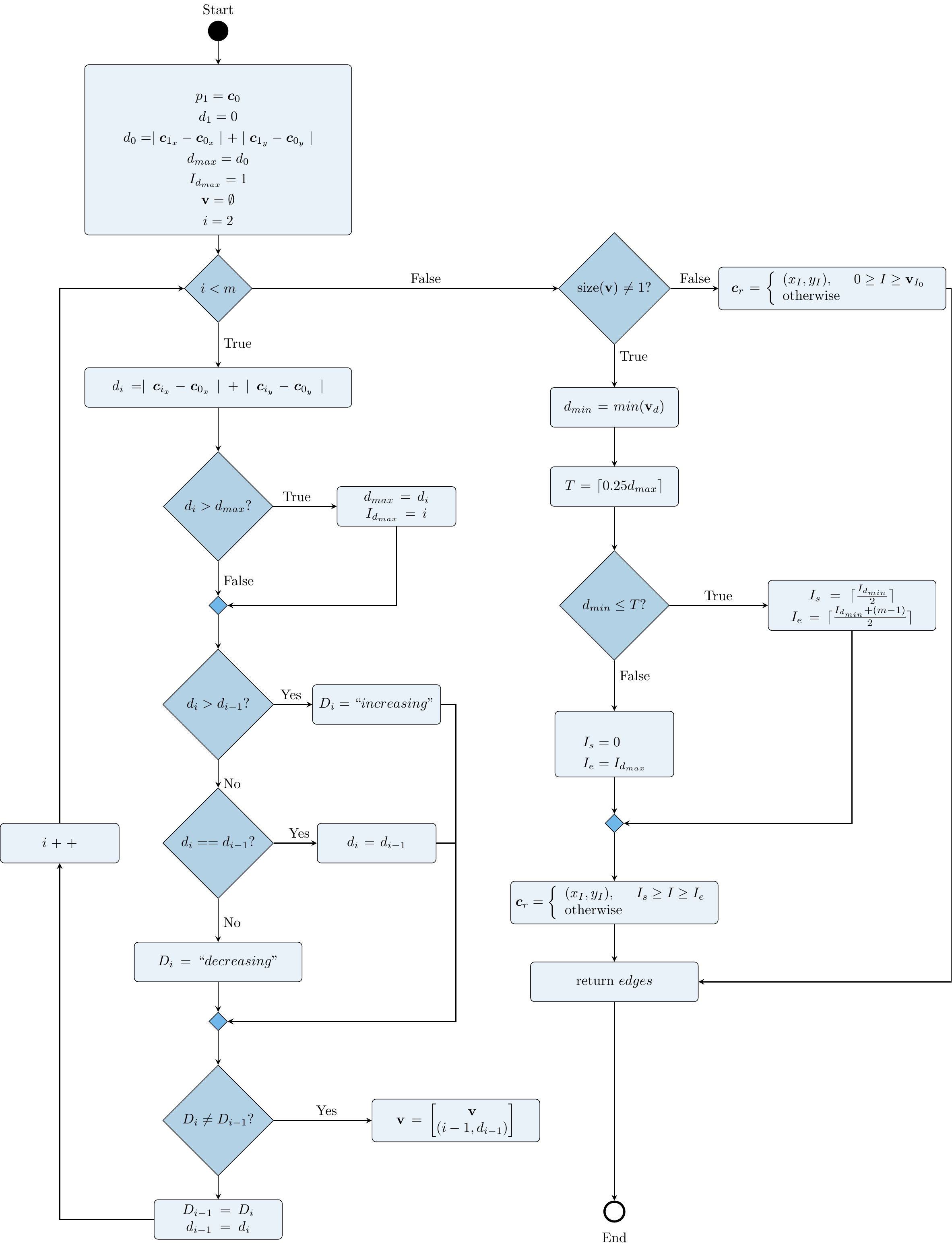}
	\end{center}
	\caption{Flowchart illustrating the contour refinement processing technique}
	\label{fig: edge extraction}
\end{figure}

The purpose of further processing the contours is to refine the contours by eliminating repeated points. The refinement of the contours is conducted due to the repeated points that complete the contour surrounding the image edge objects. Consequently, the elimination of the repeated points reduces redundant information and produces a better representation of the image edge object found in the image. A flow diagram of the contour refinement process is illustrated in Figure \ref{fig: edge extraction}. The ending index value, $I_e$ indicates the stopping point of all the unique points within the contour from the starting index point, $I_s$. All other points beyond $I_{e}$ are repeated points that forms the complete closed contour. In order to determine $I_{e}$, the maximum distance, $d_{max}$ between the contour's starting point, $p_{1}$ and all other points, $p_i$, where $i \in \{2, 3, 4, ..., m\}$ and $m$ is the total number of points in the contour, are calculated. The distance between each contour point from point $p_1$ is calculated using \eqref{eq: point distance calculation}, whilst the maximum distance is expressed in \eqref{eq: point maximum distance}.

\begin{equation}
\label{eq: point distance calculation}
d_i = |x_{p_1} - x_{p_i}| + |y_{p_1} - y_{p_i}|,
\end{equation}
where, $i \in \{2, 3, 4, ..., m \}$

\begin{equation}
\label{eq: point maximum distance}
d_{max} = max(d_1, d_2, d_3, ..., d_m)
\end{equation}

The turning-points of the closed contour is determined using $d_i$, in which it provides the direction of each point from point $p_1$. The direction of each point is referred to as the directional status, $D_i$, where $D_i \in \{``increasing",``decreasing" \}$, which is expressed in \eqref{eq: directional status}. A change in direction, $\Delta D_i$ at $D_i$ in the contour is a change in directional status between $D_{i}$ and $D_{i-1}$.

\begin{equation}
\label{eq: directional status}
D_i = \left \{
\begin{array}{lr}
``increasing", & d_{i} > d_{i-1} \\
``decreasing", & d_{i} < d_{i-1}
\end{array}
\right.
\end{equation}

The change in direction, $\Delta D_i$ is used to find the turning point with the smallest distance value, $d_{min}$. Therefore all $d_i$ at $D_i$ that are classified as a turning points are stored in vector, \textbf{v}. Vector $\boldsymbol{v}_d$ stores the index and distance values in the form of $(I, d )$, where $I$ is the index value and $d$ is the distance value. The minimum distance $d_{min}$ is calculated using \eqref{eq: minimum distance}.

\begin{equation}
\label{eq: minimum distance}
d_{min} = min(\boldsymbol{v}_d)
\end{equation}

The minimum distance, $d_{min}$ is evaluated by a threshold value, $T$ that is defined in \eqref{eq: stopping point threshold}. The evaluation determines the relative positioning of the turning point compared to $d_{max}$.

\begin{equation}
\label{eq: stopping point threshold}
T = 0.25d_{max} 
\end{equation}

The purpose of this evaluation is to determine $I_{s}$ and $I_{e}$. If $\Delta d_{min} < T$, the calculation of $I_{s}$ and $I_{e}$ are illustrated by \eqref{eq: starting index} and \eqref{eq: ending index}, respectively. Otherwise $I_{s} = 0$ and $I_e = I_{d_{max}}$, where $I_{d_{max}}$ is the index value at the $d_{max}$.

\begin{equation}
\label{eq: starting index}
I_{s} = \
\Bigg\lceil \frac{I_{d_{min}}}{2} \Bigg\rceil 
\end{equation}
where $I_{d_{min}}$ is the index of the $d_{min}$ at the turning point.

\begin{equation}
\label{eq: ending index}
I_{e} = \Bigg\lceil \frac{I_{d_{min}} + (m- 1)}{2} \Bigg\rceil
\end{equation}
where $m$ is the total number of contour points.

\subsection{Image Segmentation} \label{Section: Image Segmentation}
The X-ray images of the lower human limb consists of three distinct regions, namely, the knee, leg and foot region. In order to perform fracture detection from the X-ray image using contours (or lines), the regions are separated from one another. The separation technique is intended to isolate the detected contours within the leg region from all other regions. This is to avoid any possible confusion between the detected contours in the knee and foot region with fractured contours found in the leg region. The separation methodology is based on the frequency analysis of $0^\circ$, $45^\circ$, $90^\circ$ and $135^\circ$ gradients. The gradients are extracted from the detected contours. The gradients are determined by calculating the gradient between adjacent contour points. The frequency analysis evaluates the extracted gradients at its associated vertical positioning in the X-ray image. 

\begin{figure}[ht!]
	\centering
	\subfigure[Extracted Contours with Foot Region Present]{\label{fig: footContourPresent}\includegraphics[width=0.49\textwidth]{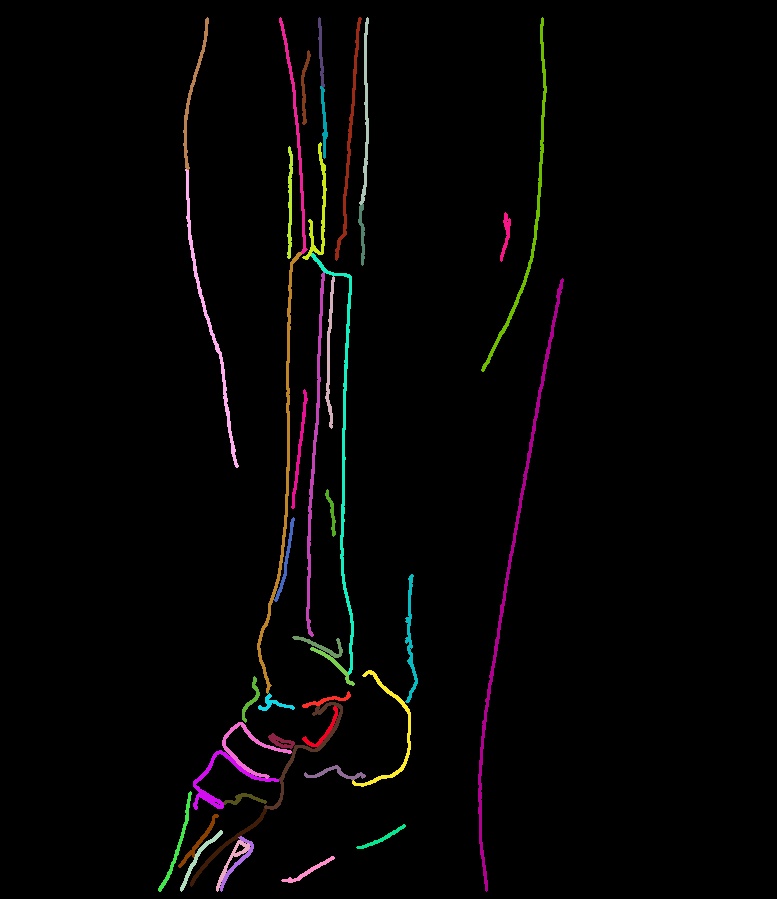}}
	\subfigure[Extracted Contours with Knee Region Present]{\label{fig: extracted contours}\includegraphics[width=0.277\textwidth]{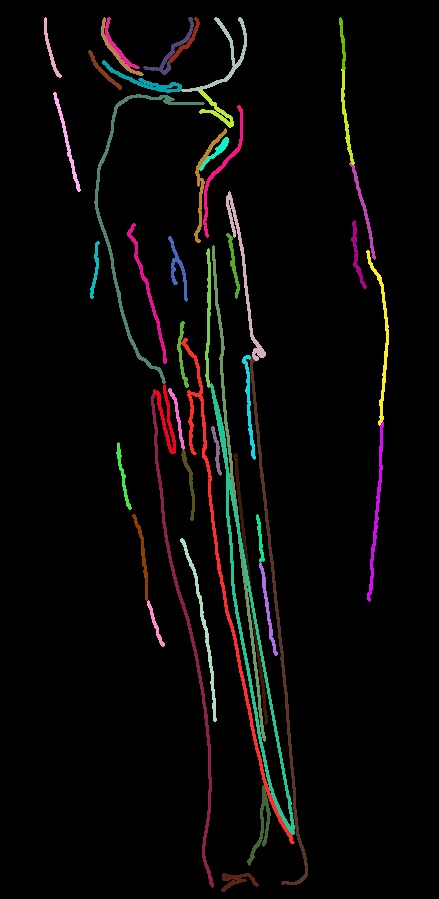}}
	\caption{Images illustrating the extracted contours in the knee and foot region} 
	\label{fig: knee and foot contours}
\end{figure}

The interpretation is conducted using the vertical positions of the image due to the positioning of the three regions. Moreover, the three regions are separated by slicing the image horizontally to achieve three distinct areas. The operation is performed through recognizing patterns in the extracted gradients. The gradients provide more information about the knee and foot region positioning within the image. Therefore, the contours within the knee and foot region have a more horizontal positioning compared to the contours within the leg region. This is illustrated in Figure \ref{fig: knee and foot contours}. As a result of the increased number of horizontal contour placements, there is a high density of $0^\circ$ gradients at particular regions within the image. In consequence, the image segmentation employs a frequency analysis technique to identify the areas of high $0^\circ$ gradient density along the vertical direction of the image to create three distinct regions.

\begin{figure}[ht!]
	\centering
	\includegraphics[scale=0.7]{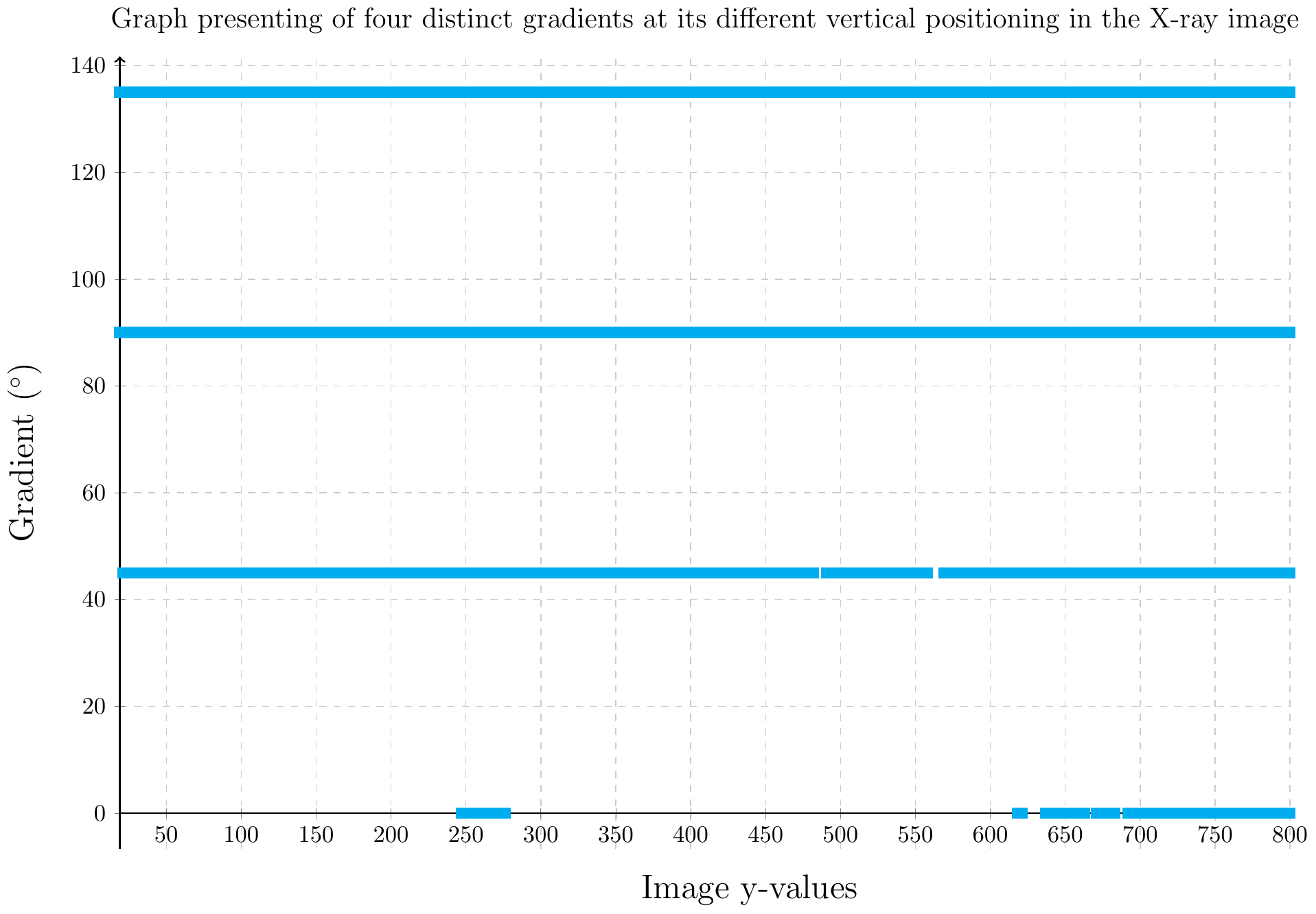}
	\label{fig:scatter_plot1}
	\caption{Density graph for each unique gradient against the image y-values}
\end{figure}

\begin{figure}[ht!]
	\centering
	\includegraphics[scale=0.85]{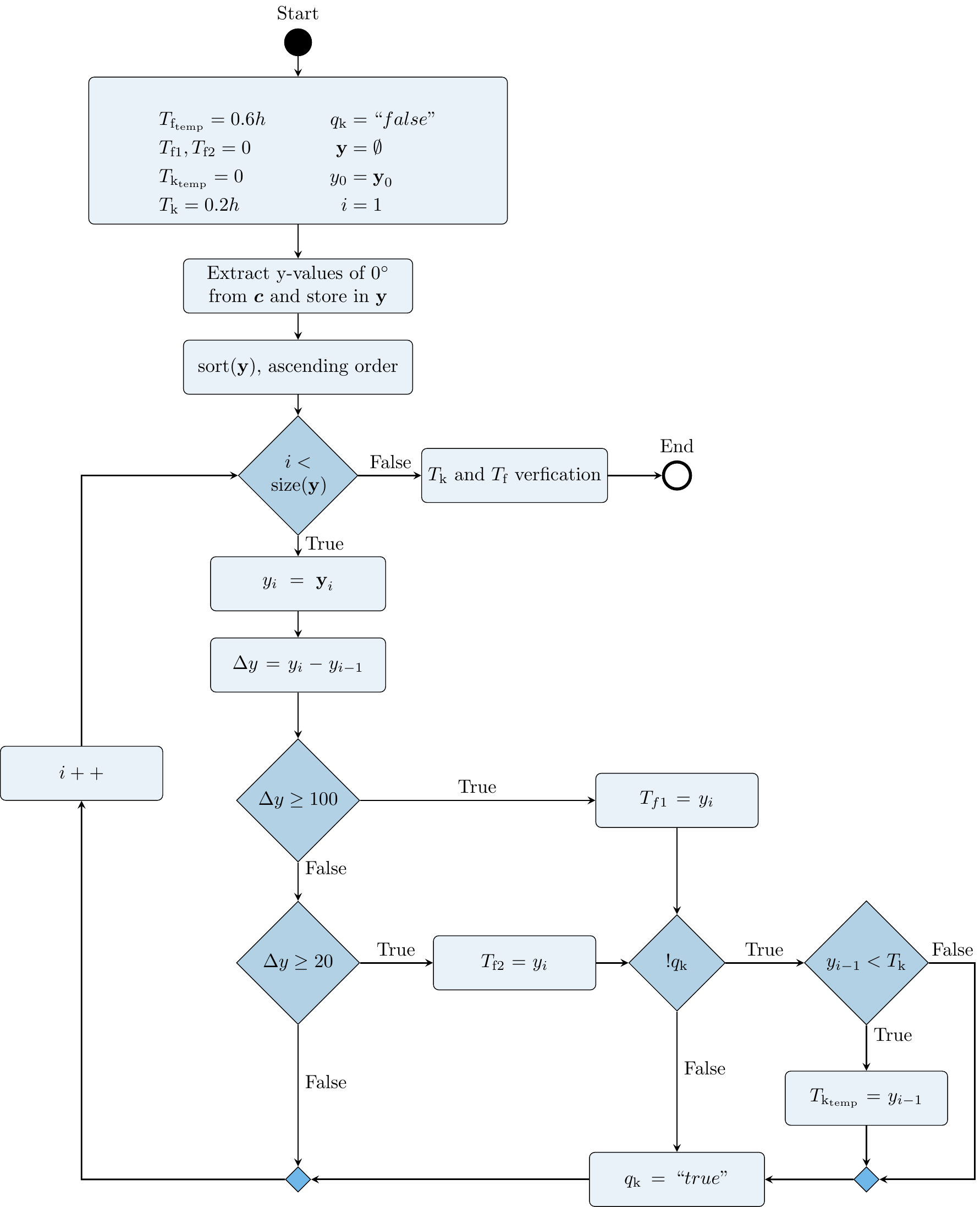}
	\caption{Flowchart illustrating the algorithm for determining the threshold values for both the knee and foot region}
	\label{fig: find temporay threshold}
\end{figure}

The results of the frequency analysis for Figure \ref{fig: footContourPresent} are presented in Figure \ref{fig:scatter_plot1}. Figure \ref{fig:scatter_plot1} clearly illustrates that there is a cluster of $0^\circ$ gradients within the foot region. This cluster of $0^\circ$ gradients is indicative of the y-value that is used to separate the foot region from the leg region. However, verification for both the knee and foot region is required. The verification of the y-value to separate both knee and foot region from the leg region is illustrated in Figure \ref{fig: find temporay threshold}.

\subsection{Region Verification} \label{Subsection: Region Verification}
The X-ray images have a portrait orientation, in which the upper portion of the lower-limb (knee region) is positioned at the top and the lower portion (foot) of the limb is at the bottom of the image. Therefore the y-values at $0^\circ$ gradient are used to verify the knee and foot region. Only the y-values at the $0^\circ$ gradients are considered, as the other three gradients do not provide information about the regions. The top left co-ordinate in $(x,y)$ of the image is $(0,0)$, whilst the bottom right co-ordinate is $(x_{\text{max}}, y_{\text{max}})$. An adaptive approach is taken to determine the threshold values for both the knee and foot regions.

First the knee threshold is obtained. The verification of the knee region is given by the temporary knee threshold, $T_{\text{k}_{\text{temp}}}$ expressed in \eqref{eq: knee_region_threshold}. The value assigned to $T_{\text{k}_{\text{temp}}}$ is based on the generalization of the knee region position within the X-ray images. The verification algorithm updates the knee threshold value, $T_\text{k}$ as it traverses through the y-values at the $0^\circ$ gradients. Figure \ref{fig: find temporay threshold} illustrates the algorithm whereby a temporary knee threshold is determined. The temporary threshold is further validated with the algorithm detailed in Algorithm \ref{alg: knee threshold verification}, whereby the temporary knee threshold is checked to determine whether the value has changed from its initial value.

\begin{equation}
\label{eq: knee_region_threshold}
T_{\text{k}_{\text{temp}}} =  0.2h
\end{equation}
where, $h$ is the height of the image.

\begin{algorithm}[ht!]
	\caption{Knee Threshold Verification Algorithm} \label{alg: knee threshold verification}
	\SetAlgoLined
	\KwData{Temporary knee threshold $T_{\text{k}_{\text{temp}}}$ value}
	\KwResult{Final knee threshold value, $T_\text{k}$}
	Assign knee threshold, $T_\text{k}$ to temporary knee threshold, $T_{\text{k}_{\text{temp}}}$ \\
	\If{$T_{\text{k}_{\text{temp}}}$ not equal to $0$}
	{
		\If{$0^\circ$ gradient cluster size $< 115$ }
		{
			$T_{\text{k}} = 0$
		}
	}
	
	\Return $T_{\text{k}}$
\end{algorithm}

For the foot region, a similar approach is employed. However, the difference is that there are two temporary foot thresholds. The first foot threshold, $T_{\text{f1}}$ is the difference between adjacent y-values, where it is used to determine large gaps between the $0^\circ$ gradient clusters. Whilst the second foot threshold, $T_{\text{f2}}$ determines the smaller gaps between the clusters. The two threshold, $T_{\text{f}1}$ and $T_{\text{f}2}$ values are checked against the a pre-determined  foot threshold, $T_{\text{f}_{\text{temp}}}$ which is expressed in \eqref{eq: y-Threshold}. The two thresholds are checked against the $T_{\text{f}_{\text{temp}}}$ to determine the final foot threshold value, $T_\text{f}$, expressed in \eqref{eq: footThreshold}. 

\begin{equation}
\label{eq: y-Threshold}
T_{\text{f}_{\text{temp}}} = 0.6h
\end{equation}

\begin{equation}
\label{eq: footThreshold}
T_{\text{f}} = 
\begin{cases}
T_{\text{f}2}, & T_{\text{f}1} < T_{\text{f}_{\text{temp}}} \\
T_{\text{f}1}, & \text{otherwise} \\
\end{cases}
\end{equation}

\begin{figure}[ht!]
	\centering
	\subfigure[Contours in the Leg Region]{\label{fig: leg region}\includegraphics[width=0.28\textwidth]{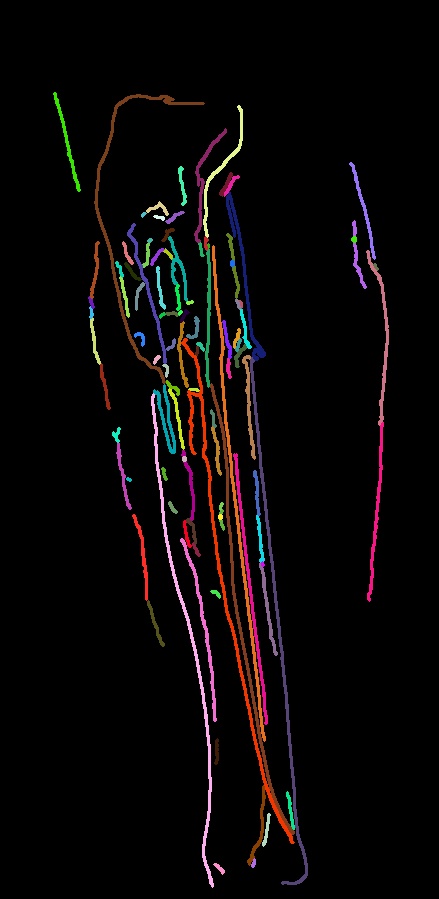}}
	\subfigure[Contours in the Foot Region]{\label{fig: foot region}\includegraphics[width=0.28\textwidth]{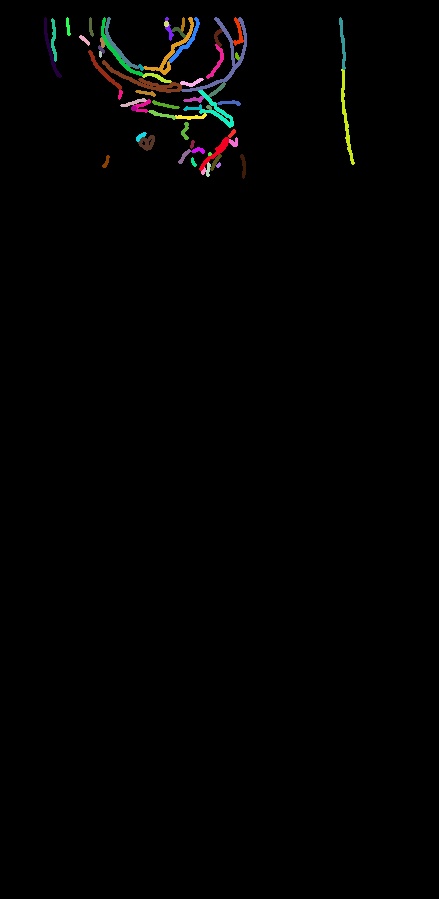}}
	\caption{Images illustrating the contours detected in the knee and leg region} 
	\label{fig: contours in leg region and knee region}
\end{figure}

\begin{figure}[ht!]
	\centering
	\subfigure[Contours in the Leg Region]{\label{fig: Original Enhanced Image}\includegraphics[width=0.49\textwidth]{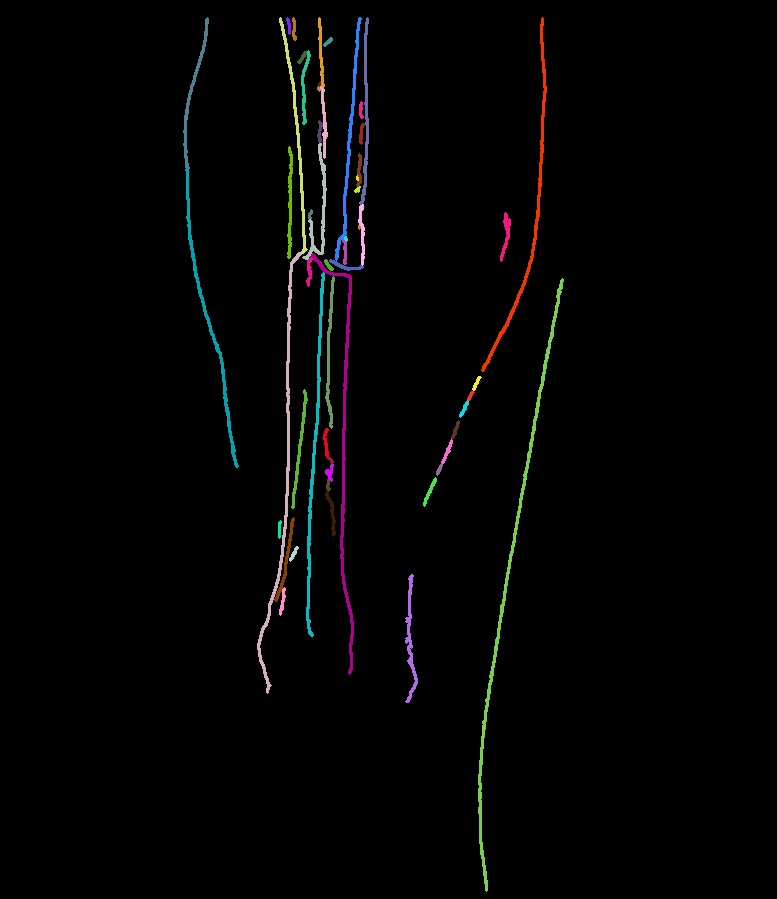}}
	\subfigure[Contours in the Foot Region]{\label{fig: extracted contours}\includegraphics[width=0.49\textwidth]{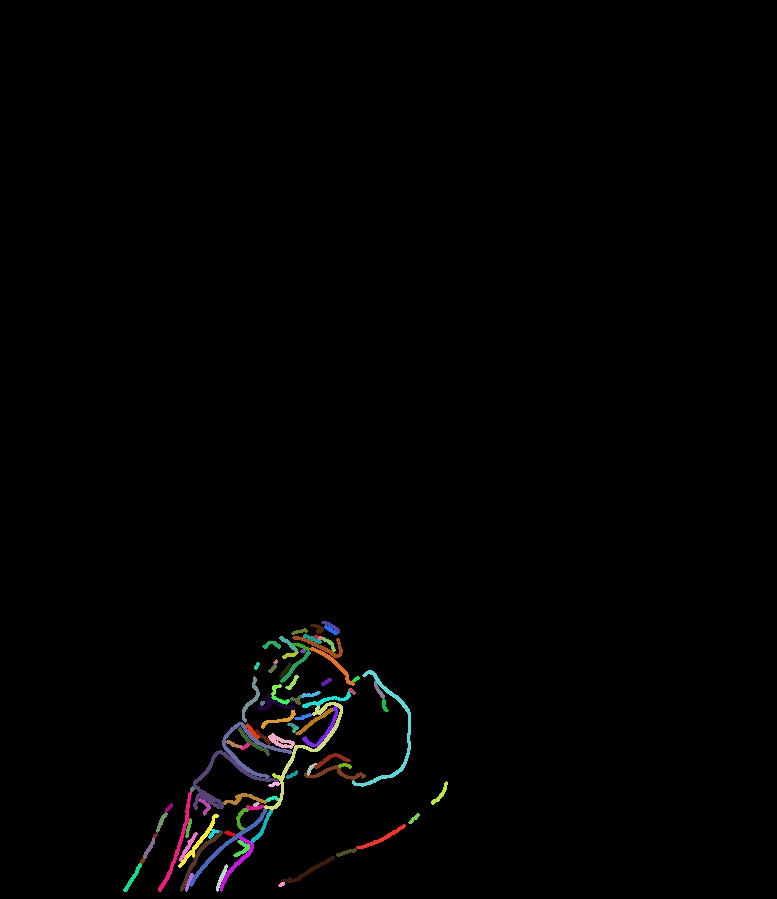}}
	\caption{Images illustrating the contours detected in leg and foot region for Figure \ref{fig: footContourPresent}} 
	\label{fig: contours in leg region and foot region}
\end{figure}

Both knee and foot thresholds are utilised to categorise contours within its respective regions: knee, leg, and foot regions. The starting or ending y-value points of the contours that are less than the knee threshold are considered part of the knee region and the contours greater than the foot threshold is considered part of the foot region, whereas all other contours fall under the leg region. The results for both knee and foot region separation from the leg region is illustrated in Figures \ref{fig: contours in leg region and knee region} and \ref{fig: contours in leg region and foot region}.

\newpage
\subsection{Contour Feature Extraction} \label{Section: Edge Feature Extraction}
There are total of 19 features extracted from each refined contour detected in the X-ray image. These features are used as inputs into the ANN for fracture and non-fracture classification. Additionally, the extracted features are a representation of the contour information, in which it summarises the information about the contour. The features are listed in Table \ref{Table: Contour Extracted Features}. Features 12 to 19 employ the histogram (frequency) analysis to determine the number of occurrences for each unique gradient derived from the contours.

\begin{table}[ht!]
	\centering
	\caption{The details of the extracted features along with its feature notation and extraction methodology for CHFB fracture detection}\label{Table: Contour Extracted Features}
	\begin{tabular}{ c p{3cm} p{1.8cm} p{1.5cm} p{7.5cm} }
		\hline
		& \textbf{Extracted Feature} & \textbf{Notation} & \textbf{Abv.} & \textbf{Extraction Methodology} \\
		\hline 
		1 & No. of Contour Points & $N_c$ & N-C & The total number of processed contour points. \\ \hline 
		2 & x start & $x_1$ & X1 & $x_1$ is the x-value of the first contour point \\ \hline
		3 & y start & $y_1$ & Y1 & $y_1$ is the y-value of the first contour point \\ \hline
		4 & x end & $x_2$ & X2 & $x_2$ is the x-value of the last contour point, where $x_2 > x_1$ \\ \hline
		5 & y end & $y_2$ & Y2 & $y_2$ is the y-value of the last contour point \\ \hline
		6 & distance & $d_t$ & DIST-T & The total distance between all adjacent points in the contour.
		\begin{equation}
		\label{eq: distance equation}
		d_\text{t} = \sum_{i=1}^{m} \sqrt{(x_{2}^{(i)} - x_{1}^{(i)})^2 + (y_{2}^{(i)} - y_{1}^{(i)})^2}
		\end{equation}
		where, $m$ is the size of the contour \\ \hline
		7 & gradient & $\theta_{\text{C}}$ & G & The gradient between the first and the last point in the contour. It is determined using the gradient equation expressed in \eqref{eq: gradient equation}.
		\begin{equation}
		\label{eq: gradient equation}
		\theta_\text{C} = tan^{-1}\Big(\frac{x_2 - x_1}{y_2 - y_1}\Big)
		\end{equation} \vspace{-3mm} \\ \hline
		8 & 1st Average gradient & $\theta_{\text{1stAvg}}$ & G-AVG1 & The \textit{1st-Average-gradient} feature is obtained by determining the gradient between two adjacent points within the contour.
		\begin{equation}
		\label{eq: 1st average gradient}
		\theta_{\text{1stAvg}} = \frac{\sum_{i = 1}^{n} \theta_i}{n} 
		\end{equation}
		where, $n$ is the number of gradient values and $\theta_i$ is the gradient between adjacent points in the contour.\\ \hline
		9 & 2nd Average gradient & $\theta_{\text{2ndAvg}}$ & G-AVG2 & The \textit{2nd-Average-gradient} feature is obtained using $\theta_{\text{1stAvg}}$.
		\begin{equation}
		\label{eq: 2nd average gradient}
		\theta_{\text{\text{2ndAvg}}} = \frac{\sum_{i = 1}^{k} \theta_{\text{\text{1stAvg}}}^{(i)}}{k}
		\end{equation}
		where, $k$ is the number of $\theta_{\text{1stAvg}}^{(i)}$ \\ \hline
		10 & x-Midpoint & $x_\text{M}$ & X-MID & The feature is determined by calculating the midpoint of between the x-values of each adjacent point in the contour. It is summed and divided by the total number of midpoints to obtain the average, which is expressed in \eqref{eq: x-Midpoint}.
		\begin{equation}
		\label{eq: x-Midpoint}
		x_\text{M} = \frac{\sum_{i = 1}^{m} \Big(\frac{x_{2_i} + x_{1_i}}{2}\Big)}{m}
		\end{equation}\\ \hline
	\end{tabular}
\end{table}

\begin{table}[ht!]
	\centering
	\begin{tabular}{ c p{3cm} p{1.8cm} p{1.5cm} p{7.5cm} }
		\hline
		& \textbf{Extracted Feature} & \textbf{Notation} & \textbf{Abv.} & \textbf{Extraction Methodology} \\
		\hline 
		11 & y-Midpoint & $y_\text{M}$ & Y-MID & The \textit{y-Midpoint} feature is similar to the \textit{x-Midpoint} feature. However, it is in the y-direction, which is expressed in \eqref{eq: y-Midpoint}.
		\begin{equation}
		\label{eq: y-Midpoint}
		y_\text{M} = \frac{\sum_{i = 1}^{m}\Big( \frac{y_2 + y_1}{2} \Big)}{m}
		\end{equation} \\ \hline
		12 & No. of Zero Gradients & $N_{\theta = 0}$ & N-G0 & The feature is the obtained using the frequency of zero gradients between two adjacent points. \\ \hline
		13 & No. of 45 Gradients & $N_{\theta = 45}$ & N-G45 & The feature is the frequency of $45^\circ$ between two adjacent points. \\ \hline
		14 & No. of 90 Gradients & $N_{\theta = 90}$ & N-G90 & The feature is the frequency of $90^\circ$ between two adjacent points. \\ \hline
		15 & No. of 135 Gradients & $N_{\theta = 135}$ & N-G135 & The feature is the frequency of $135^\circ$ between adjacent points. \\ \hline
		16 & No. of Zero Difference Gradients & $N_{\Delta \theta = 0}$ & N-G0-DIFF & The feature is the frequency of the difference zero gradients, which is calculated using the gradient values from features 12 - 15. \\ \hline
		17 & No. of 45 Difference Gradients & $N_{\Delta \theta = 45}$ & N-G45-DIFF & The feature is the frequency of $45^\circ$ difference in gradient between two adjacent gradient points. \\ \hline
		18 & No. of 90 Difference Gradients & $N_{\Delta \theta = 90}$ & N-G90-DIFF & The feature is the frequency of $90^\circ$ difference in gradient between two adjacent gradient points. \\ \hline
		19 & No. of 135 Difference Gradients & $N_{\Delta \theta = 135}$ & N-G135-DIFF & The feature is the frequency of $135^\circ$ difference in gradient between two adjacent gradients. \\ \hline
	\end{tabular}
\end{table}

\newpage
\subsection{Contour Feature Correlation} \label{Section: Contour Feature Correlation}
\begin{figure}[ht!]
	\centering
	\includegraphics[scale=0.56]{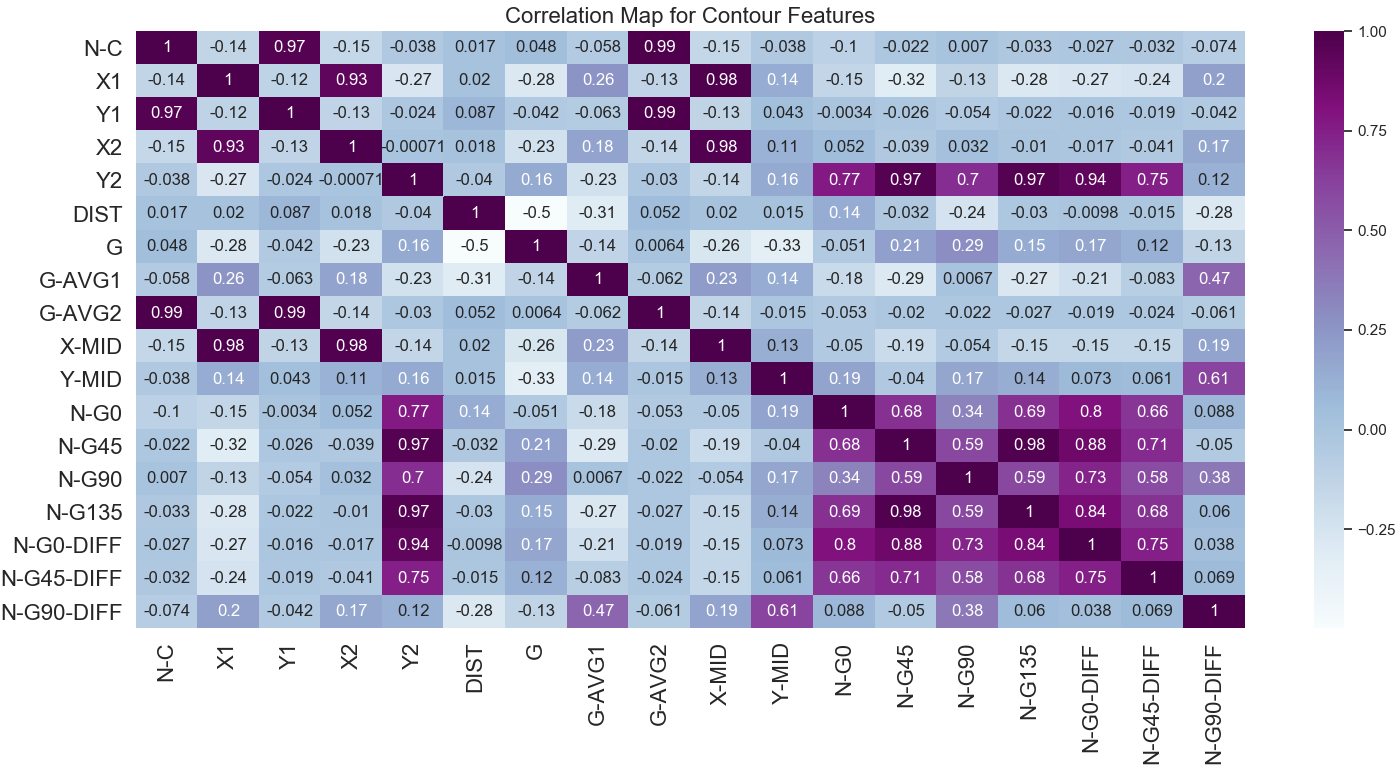}
	\caption{Correlation map illustrating the dependency between the 19 extracted contour features}
	\label{fig: contour feature correlation}
\end{figure}

Feature correlation is performed on the features extracted from the contours to investigate the dependency and independence of each feature from all other features \cite{correlation_analysis}. The correlation map is presented in Figure \ref{fig: contour feature correlation}. There are two main types of correlations, the first is strong positive correlation which is associated to the value ``1" and the second is strong negative correlation which is associated to ``-1". An independent feature has a correlation value of ``0" against all other features. Figure \ref{fig: contour feature correlation} indicates that there are some strong correlations involving features \textit{Number of contour points}, \textit{x start}, \textit{y start}, \textit{x end}, \textit{y end}, \textit{2nd Average gradient},\textit{x-Midpoint}, and \textit{Number of 45 gradients}, whilst all other features have minimal dependency on one another. Therefore, the features with strong correlation share information, which is potentially redundant.

\newpage
\subsection{Principal Component Analysis} \label{Section: Contour Principle Component Analysis}
\begin{figure}[ht!]
	\centering
	\includegraphics[scale=0.65]{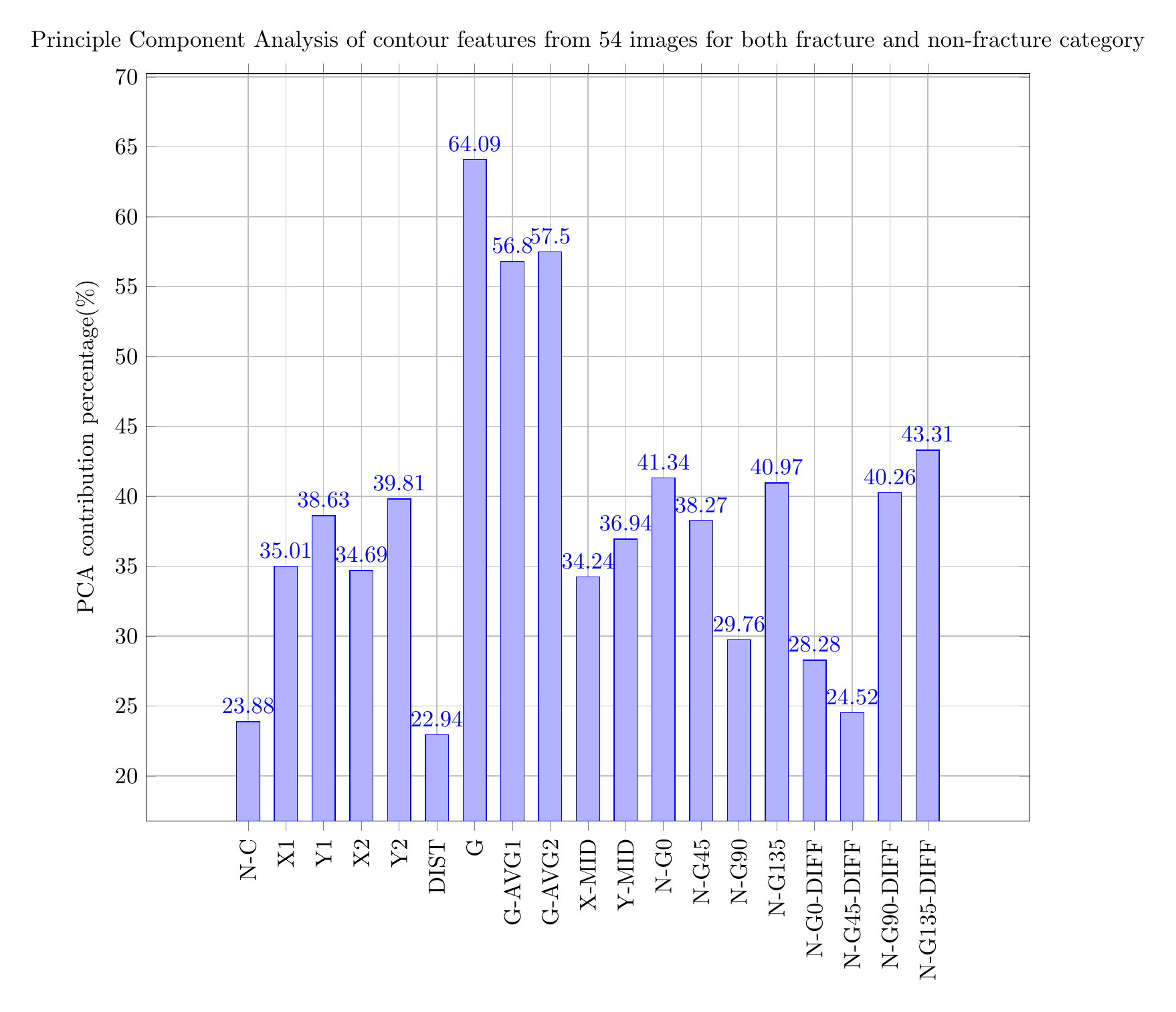}
	\caption{Graph illustrating the results of the PCA for all contours, regardless of its classification}
	\label{fig: contourPCAall}
\end{figure}

\begin{figure}[ht!]
	\centering
	\includegraphics[scale=0.65]{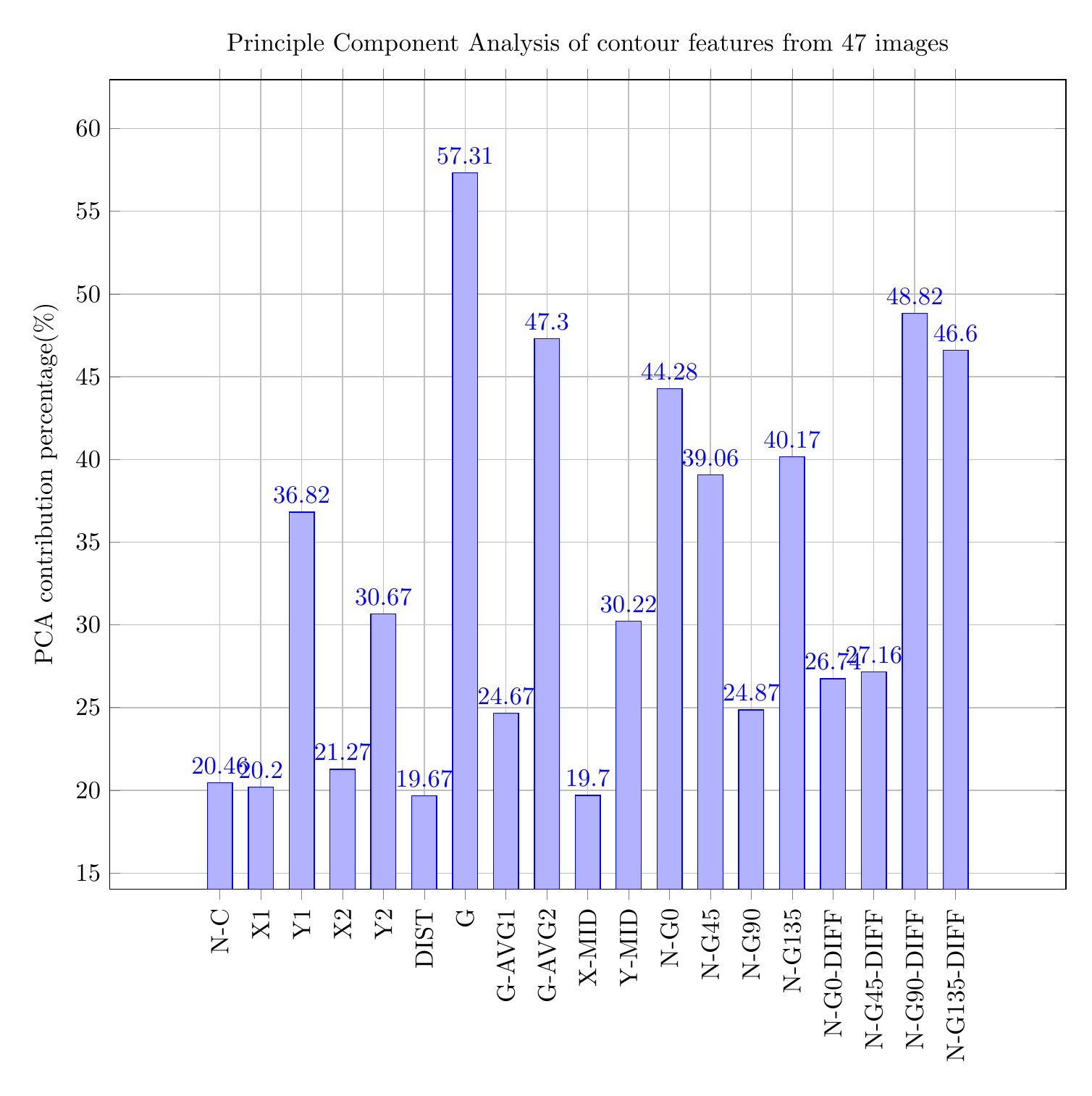}
	\caption{Graph illustrating the results of PCA for fractured contours}
	\label{fig: contourPCAaccept}
\end{figure}

\begin{figure}[ht!]
	\centering
	\includegraphics[scale=0.65]{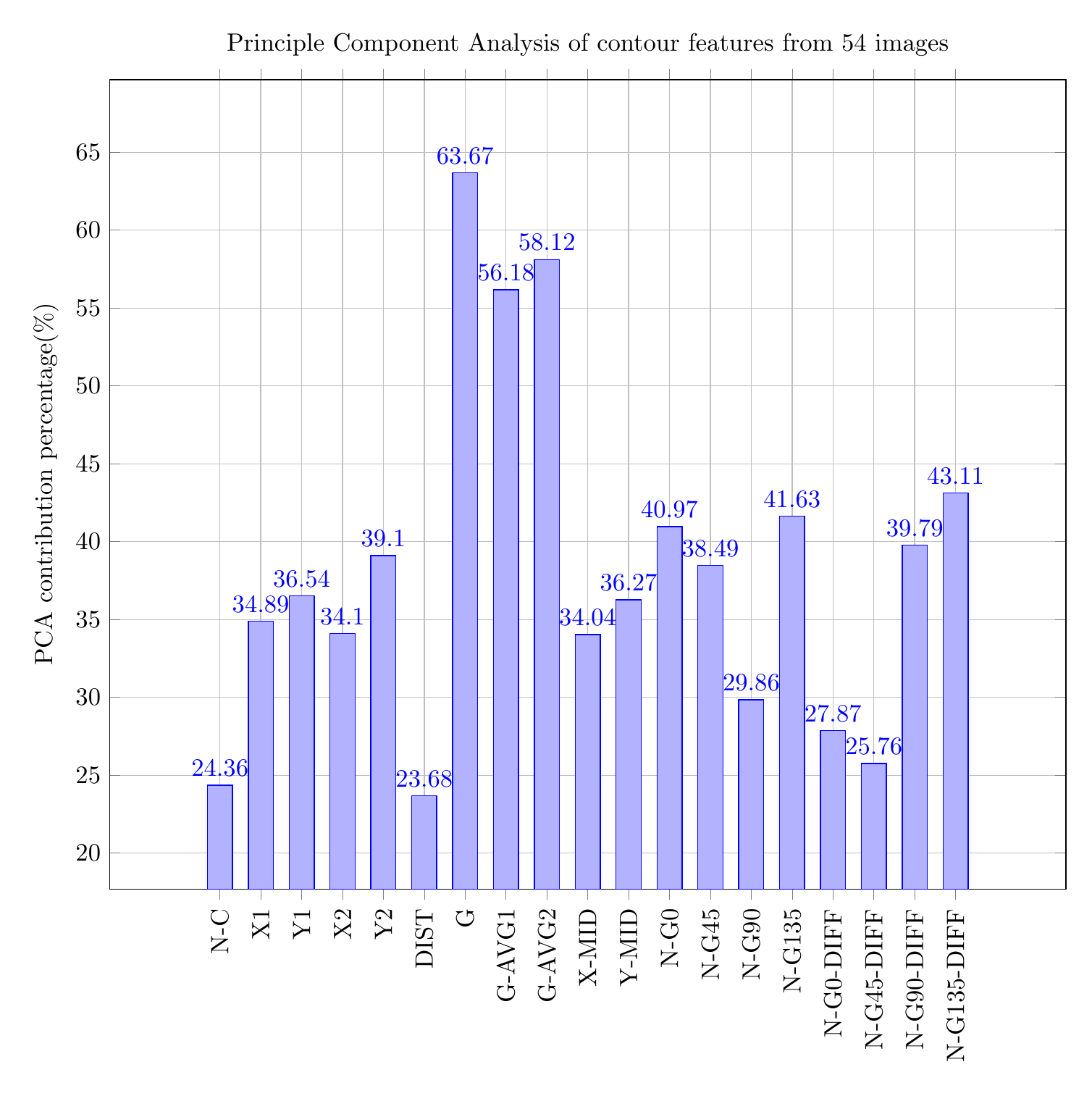}
	\caption{Graph illustrating the results of PCA for non-fractured contours}
	\label{fig: contourPCAreject}
\end{figure}

PCA is a linear dimensionality reduction technique \cite{Jolliffe2016, WOLD198737}. However, the objective of the PCA application for the Standard CHFB fracture detection scheme is to determine the dominant features within the contours that are categorized as fractures and non-fractures. There are a total of 19 features extracted for the Standard CHFB scheme. The feature contributions for a fracture is shown in Figure \ref{fig: contourPCAaccept} for 1,237 contours, whilst \ref{fig: contourPCAreject} presents the contributions of 4,576 non-fractured contours. In Figure \ref{fig: contourPCAall}, the feature contributions for 5,813 contours regardless of its given label. The overall dominant feature despite the labelling of the contours is the \textit{gradient} feature. Therefore, this is an indication that the \textit{gradient} feature holds the most varying information compared to all other features. Consequently, the \textit{gradient} feature is the dominate feature in both Figures \ref{fig: contourPCAaccept} and \ref{fig: contourPCAreject}. However, there are other sub-dominate features that differentiate a fractured contour from a non-fractured contour. For fractured contours, the sub-dominate features are \textit{Number of $90^\circ$ Difference Gradient}, followed by the \textit{1st Average gradient} feature. Whereas, for non-fractured contours the sub-dominate features are \textit{2nd Average gradient} followed by \textit{1st Average gradient}. Although both fractured and non-fractured contours have similar dominate and sub-dominate features, the contribution of each feature varies in contribution intensity.

\newpage
\subsection{Contour Data Labelling} \label{Section: Contour Data Labelling}
\begin{figure}[ht!]
	\centering
	\subfigure[Fracture Contour Selections]{\label{fig: Original Enhanced Image}\includegraphics[width=0.49\textwidth]{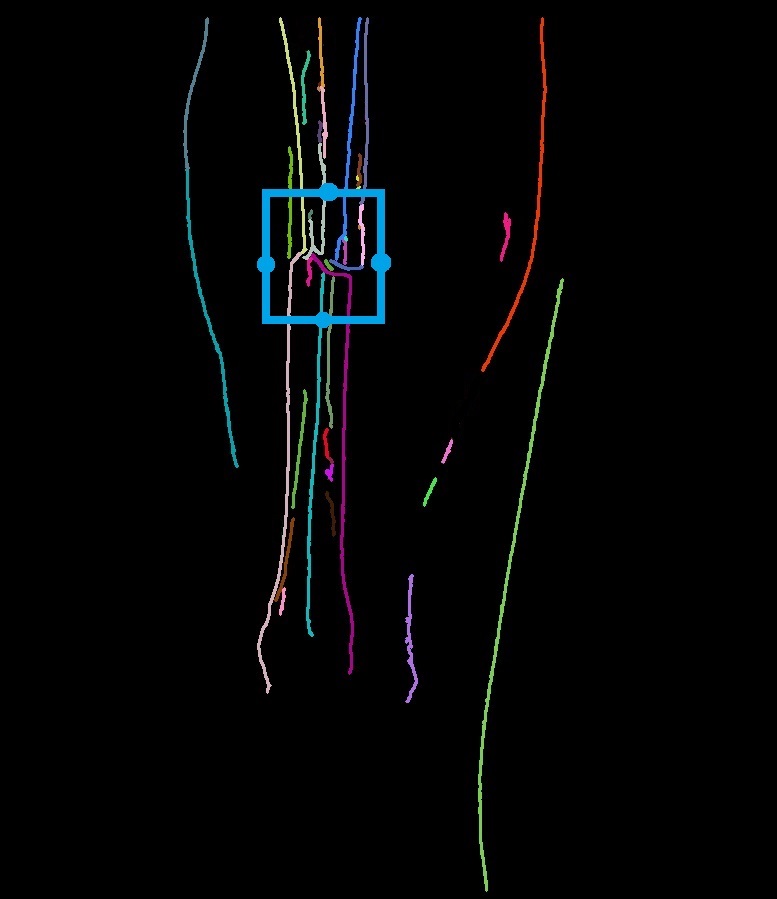}}
	\subfigure[Highlighted Fractured Contours]{\label{fig: selected contours}\includegraphics[width=0.49\textwidth]{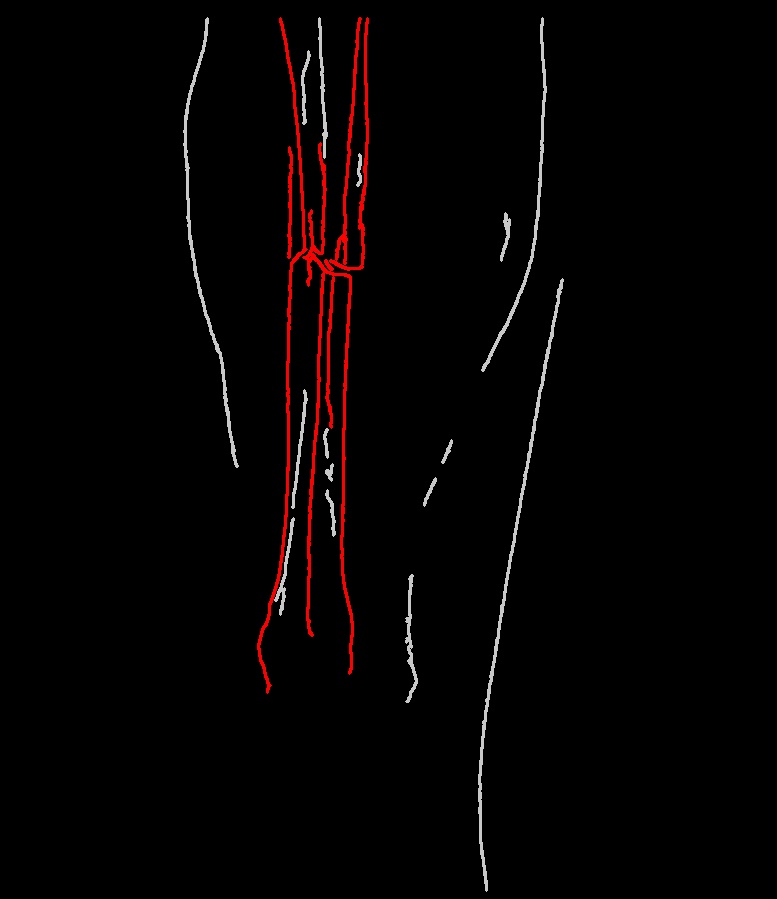}}
	\caption{Images illustrating the labelling of contours through the area selection approach} 
	\label{fig: contours data labelling}
\end{figure}

The contour data labelling employs the same approach as the data labelling methodology used in the Standard and ADPO schemes. The labelling is performed by a medical professional through a GUI, whereby the user is presented with the detected contours. The fracture region is selected by the user. An area selection approach is adopted as the individual contour labelling approach introduces difficulty in identifying whether it is fractured and non-fractured. This is due to the lack of information provided by individual contours as there is little to no context about the contour relative to its neighbours. The fractures and non-fractures are easily identified as a group since neighbouring contours provide more context about the contour within the image. Consequently, an area selection approach reduces the time consumption spent on labelling the contour data individually. A contour is labelled as a fracture if either its starting or ending point is found within the selected area. The area approach has its negatives, in that contours that are not considered fractured, but are in the selected fractured area are mislabelled as fractured. Therefore, this hinders the accuracy of the ANN. The area labelling approach for contours is illustrated in Figure \ref{fig: contours data labelling}.

\newpage
\subsection{Neural Network Architecture} \label{Section: CHFB Neural Network Architecture} 
\begin{figure}[ht!]
	\begin{center}
		\includegraphics[scale=0.9]{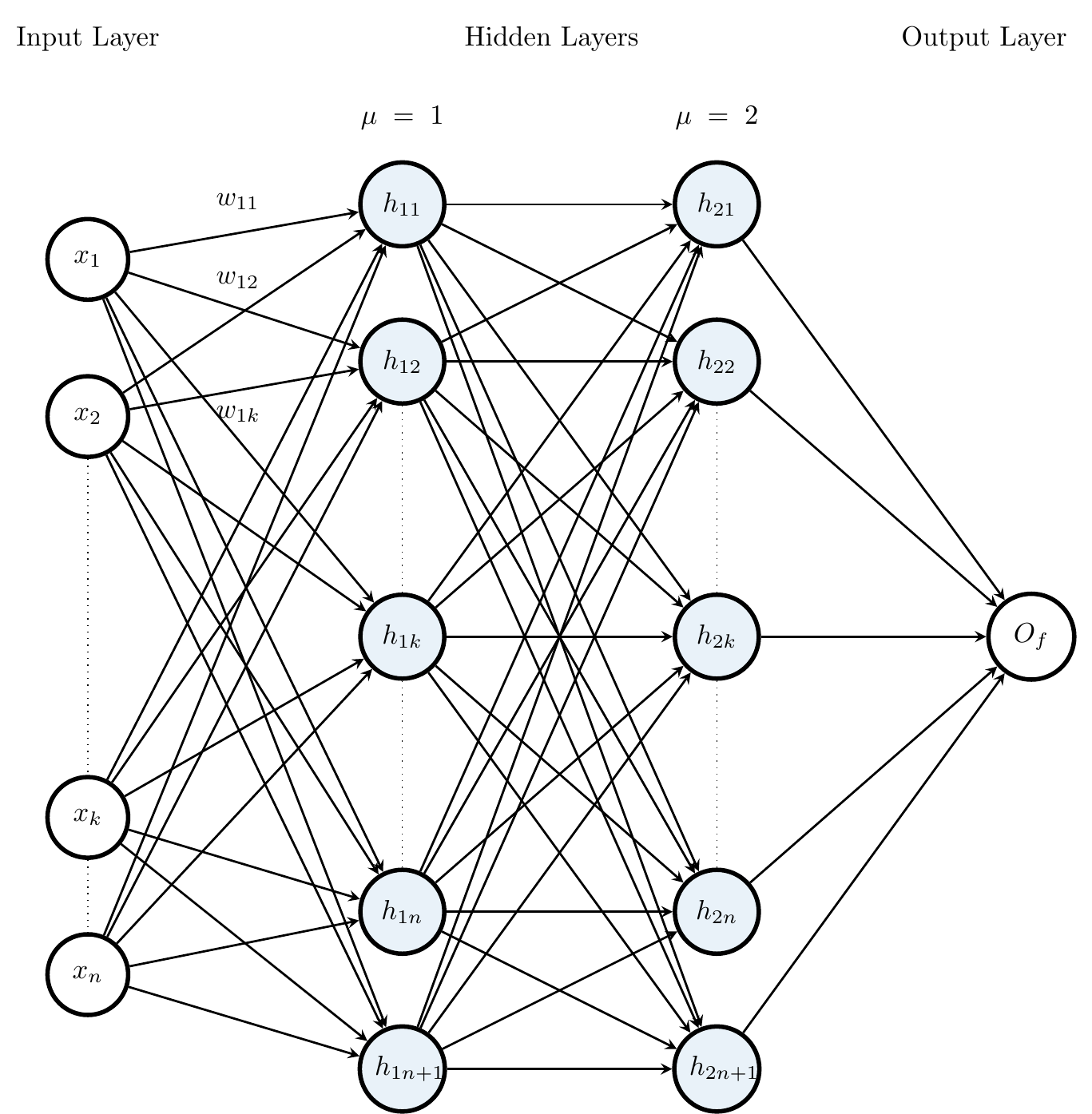}
	\end{center}
	\caption{Diagram showing the artificial neural network architecture for CHFB fracture detection scheme, where $\mu$ is indicative of the hidden layers.}
	\label{fig: line neural netwrok setup}
\end{figure}

The ANN architecture consists of four layers: one input and output layer, and two hidden layers. This a deep feed-forward neural network. The difference is the number of nodes within each layer of the ANN, except for the output layer. This is due to the additional number of features extracted from the contours. Therefore, the value assigned to $n$ in Figure \ref{fig: line neural netwrok setup} is $n = 22$. The additional three nodes are for the knee, leg and foot regional classification.

\newpage
\section{Results Analysis} \label{Section: CHFB Neural Network Results}
\begin{figure}[ht!]
	\centering
	\includegraphics[scale=0.76]{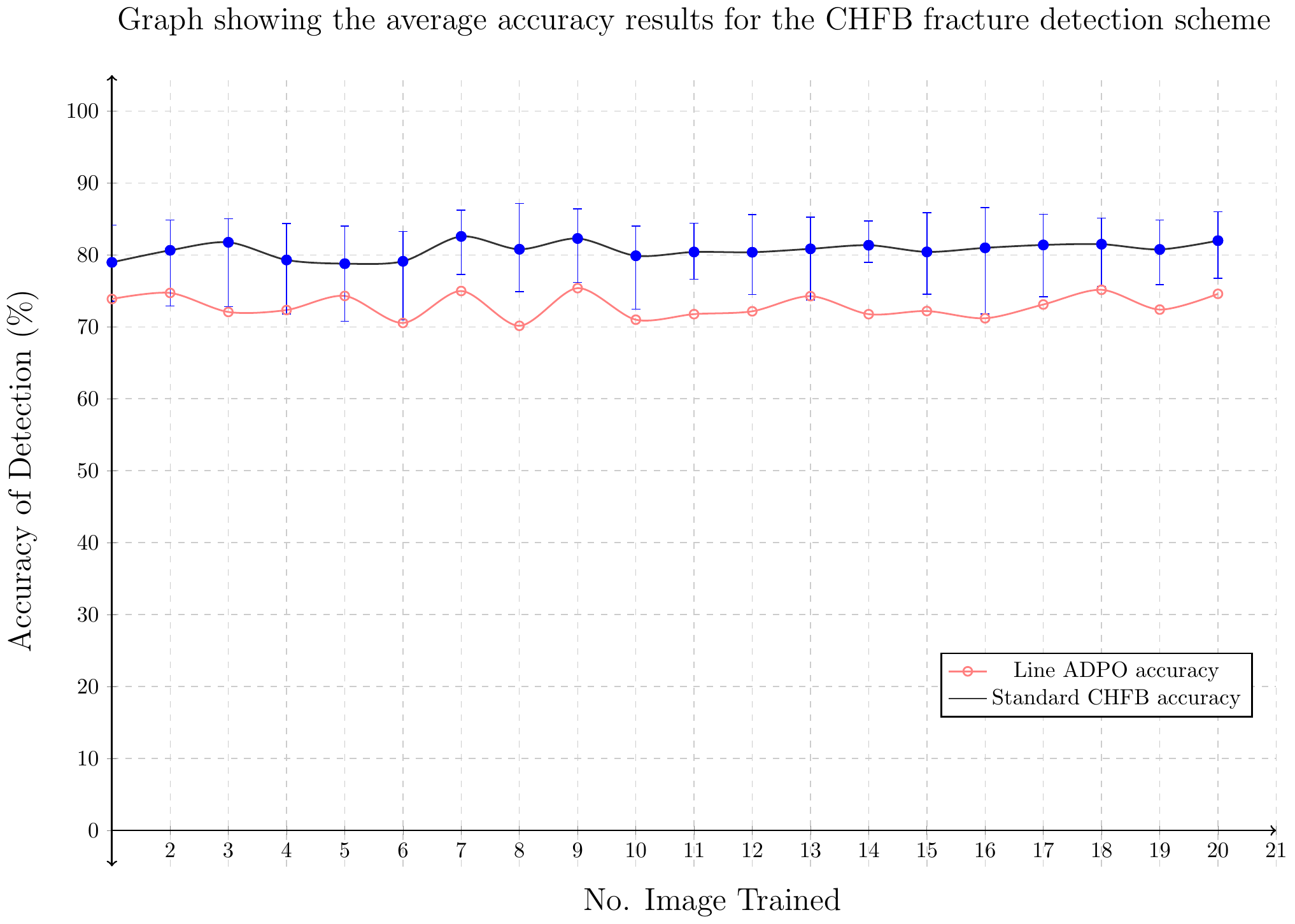}
	\caption{Graph illustrating the average accuracy for 20 cases over 10 simulations for the Standard CHFB fracture detection scheme}
	\label{fig: contour accuracy graph with error bars}
\end{figure}

The Standard CHFB fracture detection scheme is evaluated using two different evaluations. The first evaluation evaluates the scheme on a system performance basis, whilst the second evaluation evaluates the performance of the ANN. The first evaluation employs the use of contours with image context. A total of 20 images along with their associated contours are randomly selected for training the ANN. Thus, there is a total of 20 cases for the evaluation of the system's performance. Each case has 10 simulations, to adequately evaluate the system, as images are selected randomly for each case from a pool of 29 images. There is an average of 112 contours per image with a ratio of 1:3.70 for fractured to non-fractured contours. Each case is evaluated with a total number of 1,720 contours obtained from 22 images. Of the 1,720 contours, 287 contours are fractured and 1,433 are non-fractured contours. The results for each case is illustrated in Table \ref{Table: Contour Neural Network Accuracy}. The consistency of the detection accuracy is graphically presented in Figure \ref{fig: contour accuracy graph with error bars}. The accuracy of the system evaluation ranges from 74.3\% to 85.17\%. Thus, yielding an average accuracy of 80.7\%.

\begin{figure}[ht!]
	\centering
	\includegraphics[scale=0.7]{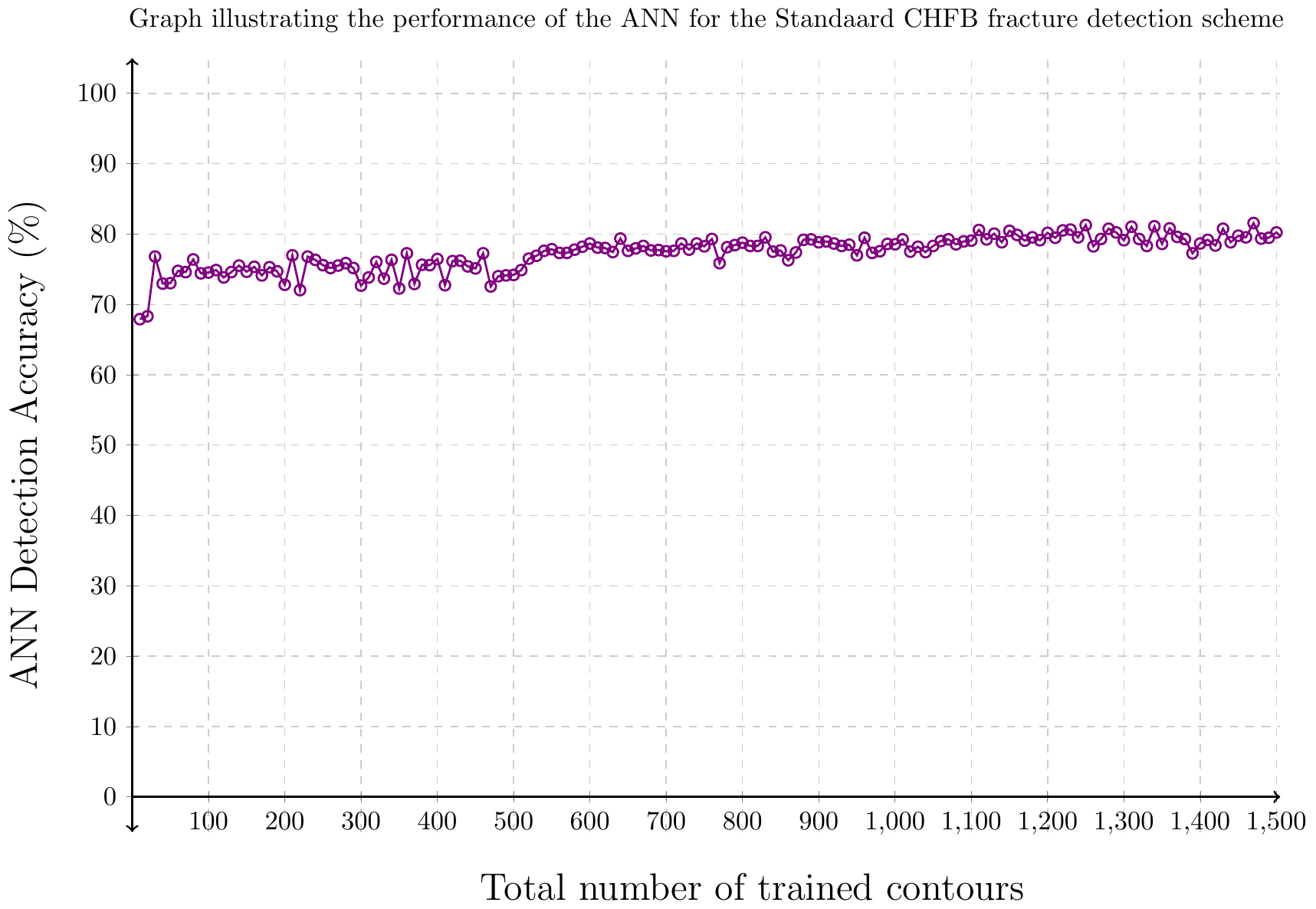}
	\caption{Graph illustrating the performance of the ANN by training it with an equal number of fractured and non-fractured contours for the Standard CHFB fracture detection scheme}
	\label{fig: contour accuracy}
\end{figure}

The second evaluation selects random contours, which do not have any image context that the contour is detected from. There are an equal number of fractured to non-fractured contours used to train the ANN for each case. Therefore, the training data consists of 50\% fractured contours and 50\% non-fractured contours. There are a total of 150 cases, in which the fractured and non-fractured contours are grouped in multiples of 5's for each case. Hence, a total number of 1,500 lines are used to train the ANN. The results are presented in Figure \ref{fig: contour accuracy}. The results shows that the accuracy of the ANN ranges from 78\% to 83\%.

\begin{table}[ht!]
	\centering
	\caption{The results of the system's minimum, average and maximum accuracies for 20 cases over 10 simulations for the Standard CHFB fracture detection scheme}
	\label{Table: Contour Neural Network Accuracy}
	\begin{tabular}{c c c c}
		\hline
		\textbf{No. Trained Images} & \textbf{Min Accuracy (\%)} & \textbf{Average Accuracy (\%)} & \textbf{Max Accuracy (\%)} \\
		\hline 
		1 & 73.547 & 78.9593 & 84.128 \\
		2 & 72.907 & 80.6338 & 84.826 \\
		3 & 72.791 & 81.75 & 85.00 \\
		4 & 71.802 & 79.2906 & 84.36 \\
		5 & 70.756 & 78.7792 & 84.012 \\
		6 & 70.93 & 79.1047 & 83.256 \\
		7 & 77.267 & 82.5638 & 86.221 \\
		8 & 74.884 & 80.7792 & 87.151 \\
		9 & 76.163 & 82.2791 & 86.395 \\
		10 & 72.442 & 79.8721 & 84.012 \\
		11 & 76.628 & 80.4012 & 84.419 \\
		12 & 74.477 & 80.3662 & 85.581 \\
		13 & 73.721 & 80.8374 & 85.233 \\
		14 & 78.953 & 81.343 & 84.709 \\
		15 & 74.535 & 80.4128 & 85.872 \\
		16 & 71.802 & 80.9767 & 86.57 \\	
		17 & 74.186 & 81.3779 & 85.64 \\
		18 & 75.64 & 81.4826 & 85.116 \\
		19 & 75.872 & 80.7559 & 84.826 \\
		20 & 76.744 & 81.9769 & 85.988 \\
		\hline
	\end{tabular}
\end{table}

\newpage
The ROC graph is way of evaluating binary classification systems through sensitivity and specificity \cite{delong1988comparing, hajian2013receiver}. The sensitivity and specificity is calculated using \eqref{eq: sensitivity} and \eqref{eq: specificity}, respectively.

\begin{equation}
\label{eq: sensitivity}
sensitvity = \frac{\text{TP}}{\text{TP} + \text{FN}}
\end{equation} 

\begin{equation}
\label{eq: specificity}
specificity = \frac{\text{FP}}{\text{FP} + \text{TN}}
\end{equation}

\begin{figure}[ht!]
	\centering
	\includegraphics[scale=0.71]{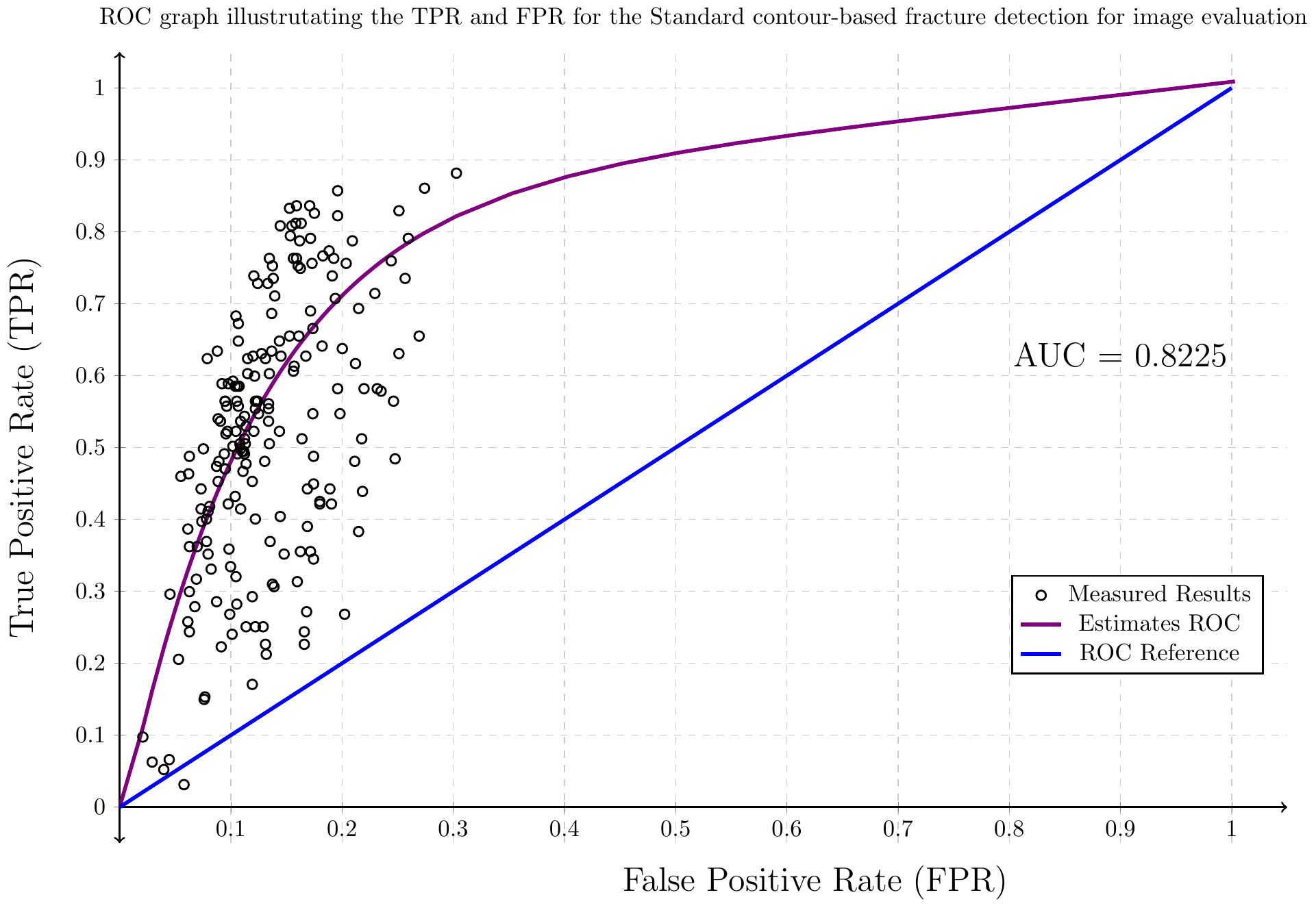}
	\caption{Figure illustrating ROC curve for the standard CHFB fracture detection}
	\label{fig: contour ROC}
\end{figure}

The ROC graph for the Standard CHFB fracture detection is presented in Figure \ref{fig: contour ROC}, in which it has a AUC value of 0.8225. The ROC graph is an indication of the system's sensitivity to detect true positives. A higher AUC value indicates better sensitivity detection, as the ideal AUC value is 1. Thus, the AUC value of 0.8225 is an improvement from the line-based fracture detection schemes. Consequently, the system reaches maximum sensitivity at 0.3 FPR from the measured results.

\newpage
\section{Data Improvement}
\begin{figure}[ht!]
	\centering
	\subfigure[Original Extracted Contours]{\label{fig: Original Extracted Contours}\includegraphics[width=0.49\textwidth]{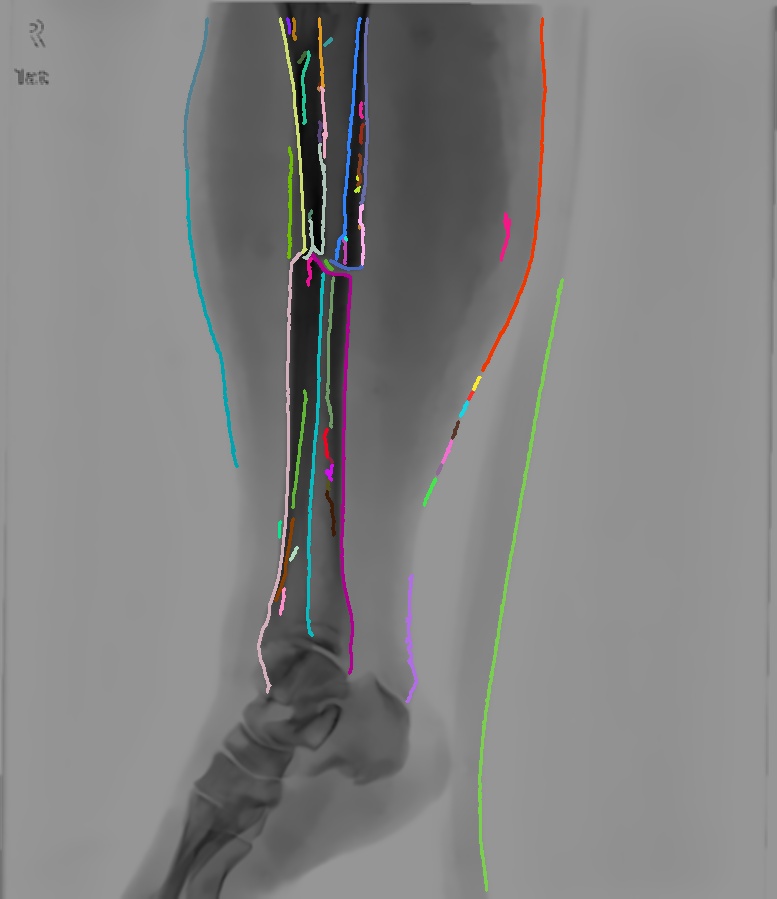}}
	\subfigure[Bone Contours Only]{\label{fig: Bone Contours Only}\includegraphics[width=0.49\textwidth]{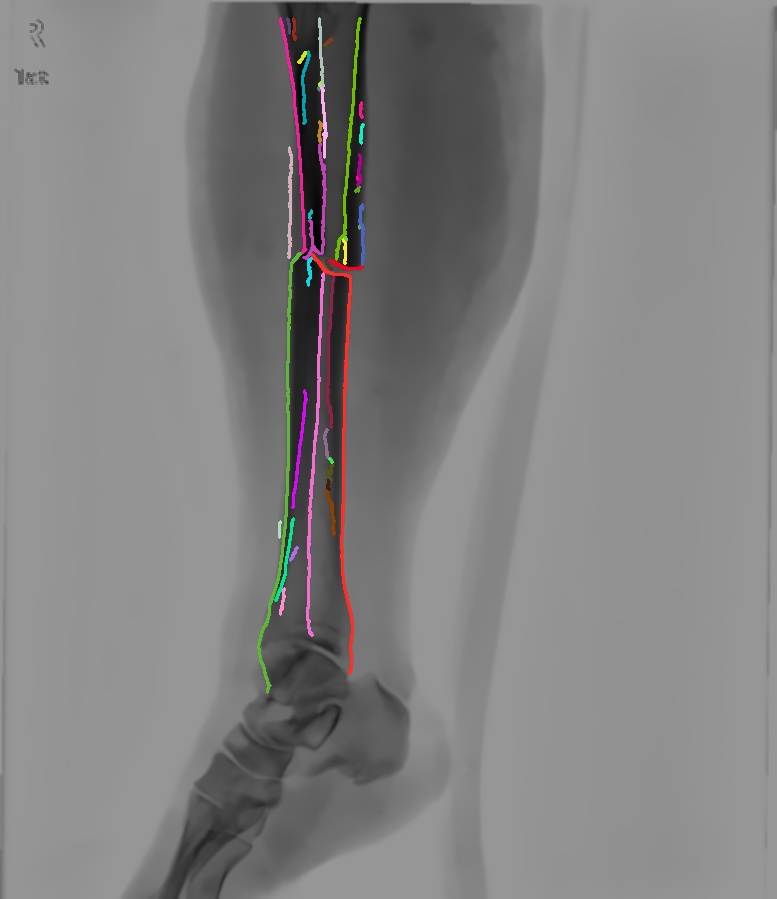}}
	\caption{Images illustrating the isolation surrounding flesh contours from the leg-bone contours} 
	\label{fig: contour data improvement}
\end{figure}

\begin{figure}[ht!]
	\centering
	\includegraphics[scale=0.75]{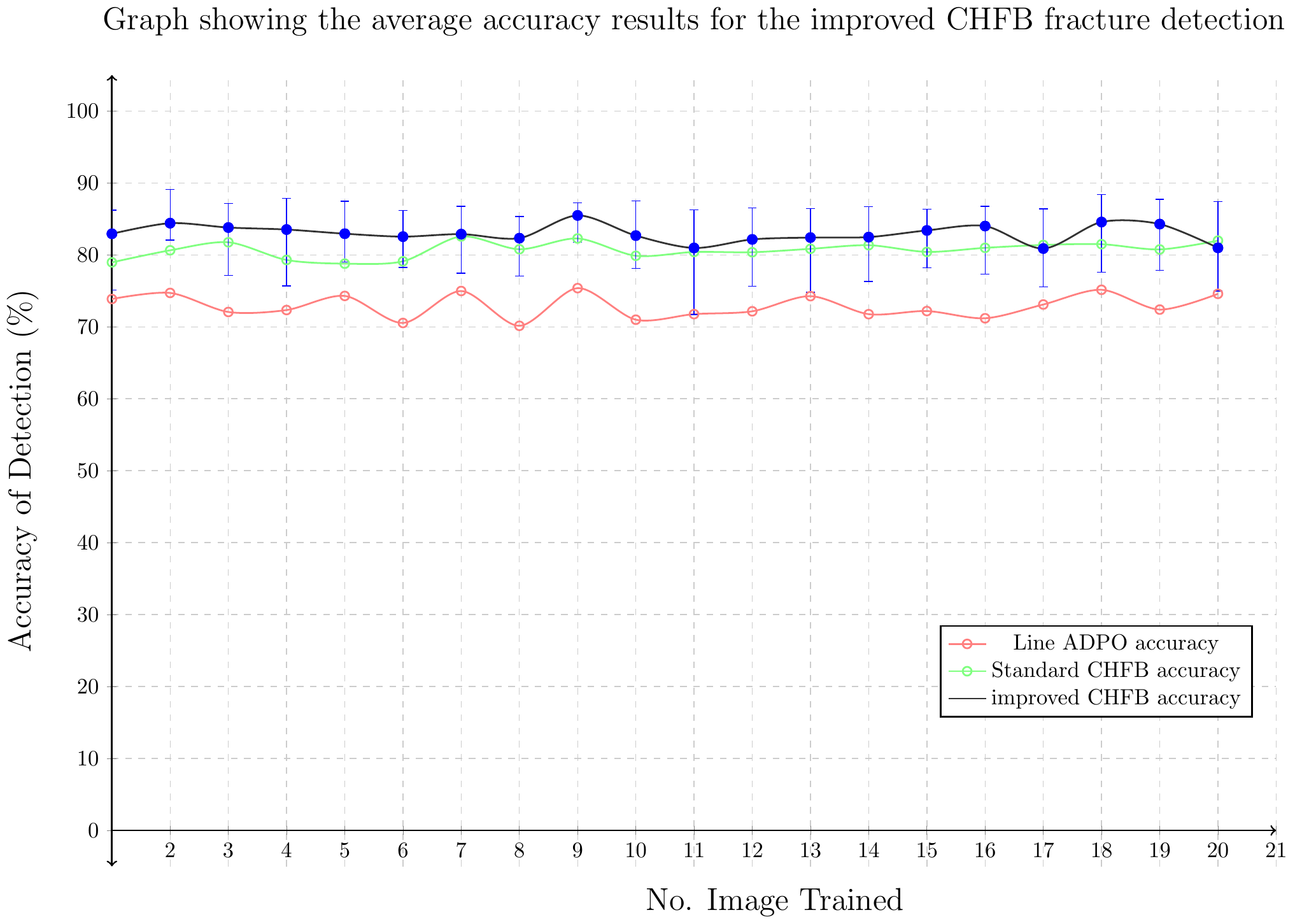}
	\caption{Graph illustrating the average accuracy for 20 cases over 10 simulations for the improved CHFB fracture detection scheme}
	\label{fig: improved contour accuracy graph with error bars}
\end{figure}

The Standard CHBF fractured detection scheme is improved by applying the x-value analysis on the starting and ending points of the contours to isolate the surrounding flesh contours from the leg-bone contours. This is illustrated in Figure \ref{fig: contour data improvement}. The isolated flesh contours are automatically classified as non-fractures. This minimises the training complexity of the ANN. Additionally, the testing data set that evaluates the ANN, does not consist of the flesh contours. It is an unfair evaluation of the ANN if the testing data set includes the flesh contours as the ANN has no exposure (knowledge) about the flesh contours. Thus, the average number of contours per image is 108 and the fractured to non-fractured ratio is 1:3.99 for the improved CHFB scheme. The system evaluation of the improved CHFB scheme is detailed in Table \ref{Table: Improved Contour Neural Network Accuracy}, whereby 1,866 contours are used to evaluate the system. Of the 1,866 contours, 277 are fractured and 1,433 are non-fractured contours. The accuracy of the system ranges from 77.08\% to 87.0\%, in which it yields an average accuracy of 82.98\%. A graphical illustration of the accuracy is illustrated in Figure \ref{fig: improved contour accuracy graph with error bars}. The comparison between Figures \ref{fig: contour accuracy graph with error bars} and \ref{fig: improved contour accuracy graph with error bars} shows that there is a slight improvement in the accuracy for the improved CHFB scheme. The accuracy improved from an 80.7\% average accuracy to a 82.98\% average accuracy.

\begin{table}[ht!]
	\centering
	\caption{The results for the system's minimum, average and maximum accuracies for 20 cases over 10 simulations for the improved CHFB fracture detection scheme}
	\label{Table: Improved Contour Neural Network Accuracy}
	\begin{tabular}{c c c c }
		\hline
		\textbf{No. Trained Images} & \textbf{Min Accuracy (\%)} & \textbf{Average Accuracy (\%)} & \textbf{Max Accuracy (\%)} \\
		\hline \hline
		1 & 75.134 & 82.9314 & 86.227 \\
		2 & 82.047 & 84.4052 & 89.068 \\
		3 & 77.17 & 83.7996 & 87.138 \\
		4 & 75.67 & 83.5209 & 87.835 \\
		5 & 78.992 & 82.942 & 87.46 \\
		6 & 78.242 & 82.5348 & 86.174 \\
		7 & 77.438 & 82.8831 & 86.763 \\
		8 & 77.063 & 82.3205 & 85.316 \\
		9 & 81.726 & 85.4824 & 87.245 \\
		10 & 78.135 & 82.6796 & 87.513 \\
		11 & 71.704 & 80.9593 & 86.281 \\
		12 & 75.616 & 82.149 & 86.549 \\
		13 & 74.812 & 82.4061 & 86.442 \\
		14 & 76.313 & 82.4706 & 86.71 \\
		15 & 78.189 & 83.3976 & 86.334 \\
		16 & 77.331 & 84.0032 & 86.763 \\
		17 & 75.563 & 80.879 & 86.388 \\
		18 & 77.599 & 84.5658 & 88.371 \\
		19 & 77.867 & 84.2765 & 87.728 \\
		20 & 74.973 & 80.9754 & 87.406 \\		
		\hline
	\end{tabular}
\end{table}

\begin{figure}[ht!]
	\centering
	\includegraphics[scale=0.75]{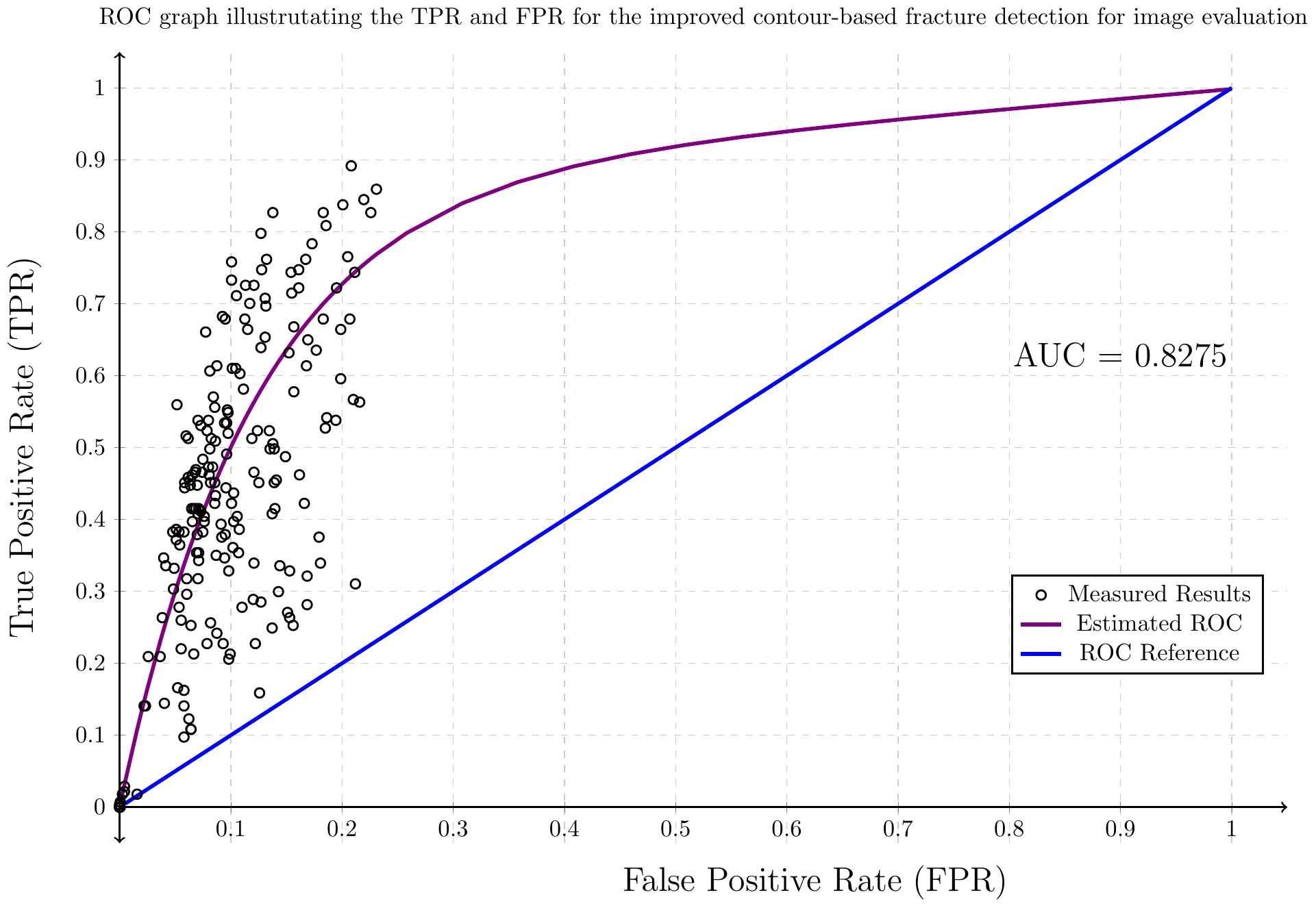}
	\caption{Figure illustrating ROC curve for the improved CHFB fracture detection}
	\label{fig: improved contour ROC}
\end{figure}

The AUC graph for the the improved CHFB scheme is presented in Figure \ref{fig: improved contour ROC}, in which it has an AUC value of 0.8275. This is a slight improvement from the Standard CHFB scheme that has an AUC value of 0.8225. Therefore the improved CHFB has a better sensitivity than the Standard CHFB scheme. Consequently, the improved CHFB scheme reaches maximum sensitivity at 0.21 FPR from the measured results.

\begin{figure}[ht!]
	\centering
	\includegraphics[scale=0.75]{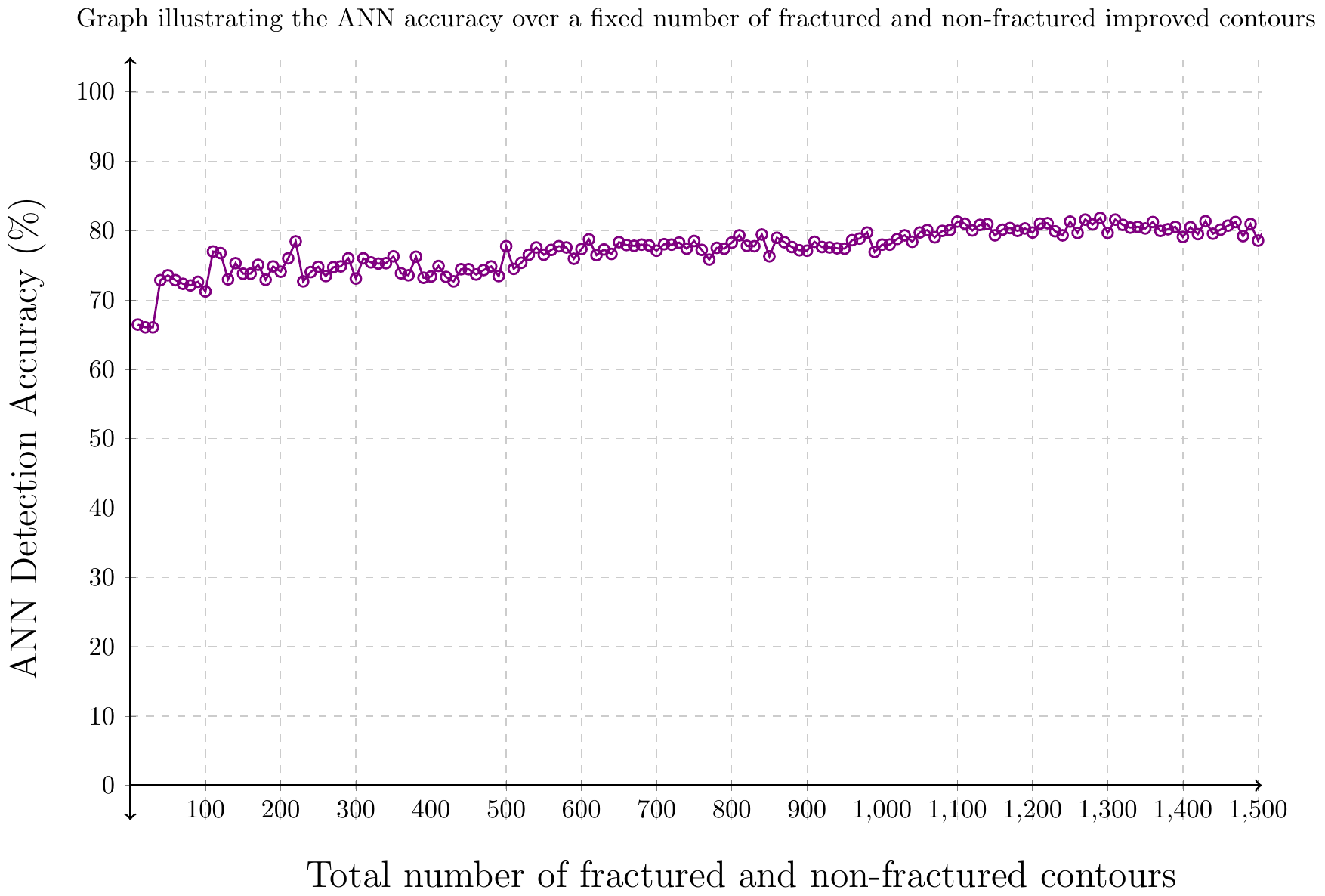}
	\caption{Graph illustrating the accuracy of the ANN for training an equal number of fixed fractured and non-fractured contours}
	\label{fig: improved contour accuracy}
\end{figure}

Figure \ref{fig: improved contour accuracy} illustrates that the ANN evaluation for the new contour data over a fixed numbers of total contours, whereby 50\% are fractured and 50\% are non-fractured contours. The figure shows that the accuracy of the ANN has a similar pattern to Figure \ref{fig: contour accuracy}, in which both figures show a gradual increase in accuracy as the total number of contours used for training increases. The ANN performance is evaluated using a fixed number of contours of 1,718 contours which are extracted from 22 images. There are 277 contours labelled as fractures, whilst there are 1,441 contours labelled as non-fractures.

\section{Clustering} \label{Section: Clustering}
\begin{figure}[ht!]
	\centering
	\subfigure[Detected Contour Fractures]{\label{fig: Detected Contour Fractures}\includegraphics[width=0.49\textwidth]{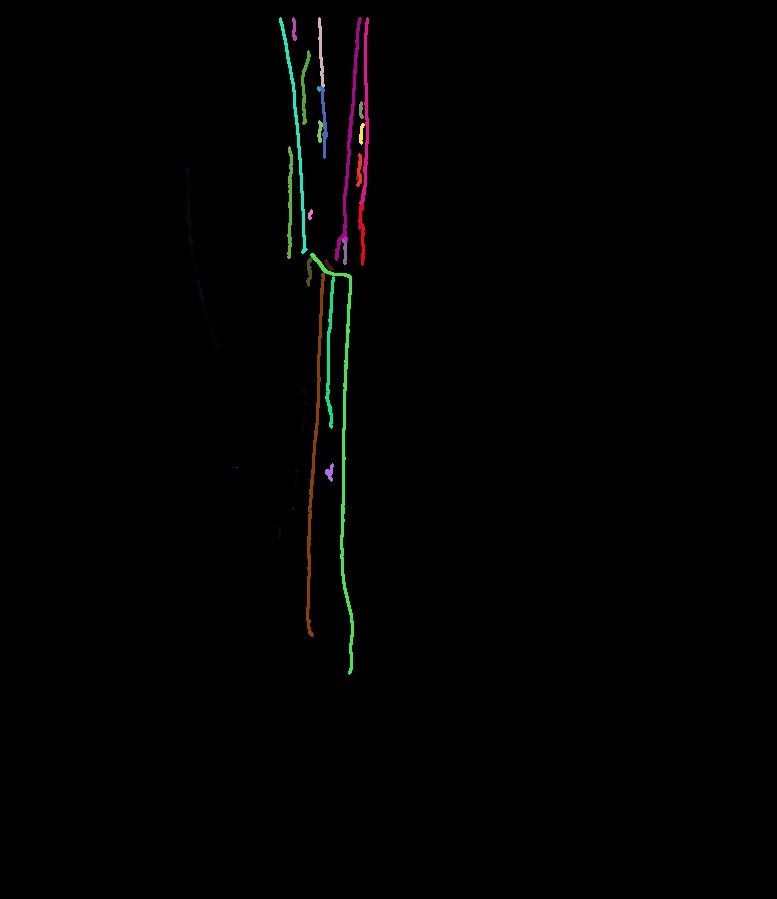}}
	\subfigure[Fracture Region]{\label{fig: Bone Contours Only}\includegraphics[width=0.49\textwidth]{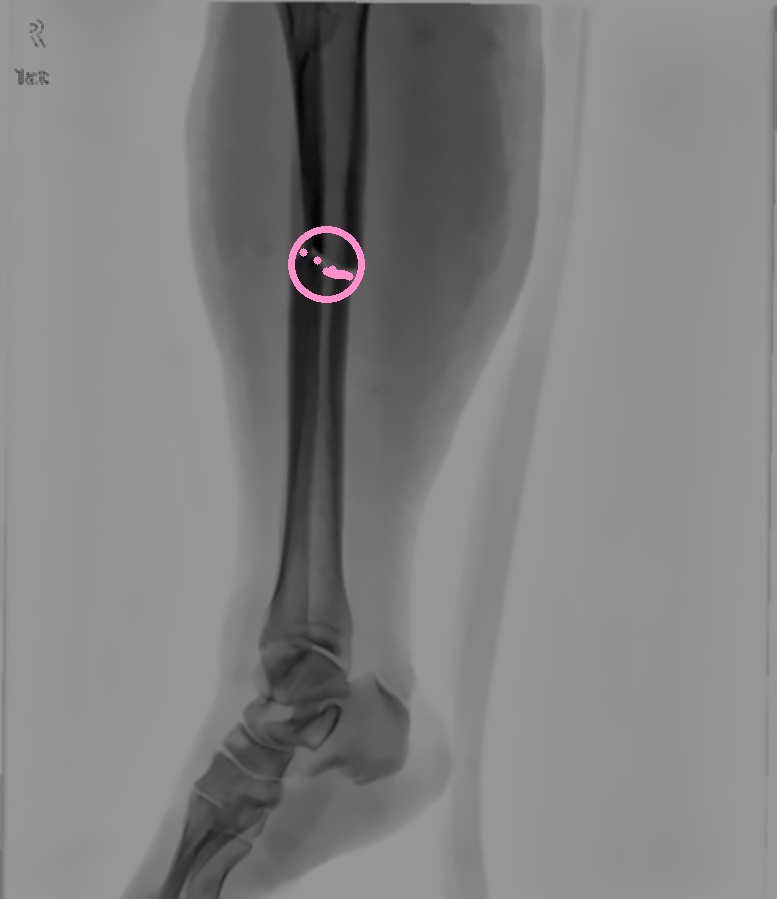}}
	\caption{Images illustrating the detection of the fractured contours and its highlighted fractured region} 
	\label{fig: contour clustering}
\end{figure}

The detection for fractured contours is performed by the contour-based fracture detection scheme, in which contours classified as fractures are illustrated to the user through the GUI. The results of the detected fractures from the improved CHFB scheme are illustrated in Figure \ref{fig: Detected Contour Fractures}. However, there are contours which extend beyond the fractured area which do not highlight the fractured region. In order to highlight the fractured region, the $0^\circ$ gradient points of the fractured contours are extracted for further processing to highlight the fractured region. The processing technique adopted is the hierarchical clustering technique.

Hierarchical clustering is the gathering of points based on the distance between the points itself to form clusters. Smaller clusters are gathered to form larger clusters, hence creating a hierarchy of clusters \cite{noauthor_hierarchical_nodate, johnson_hierarchical_1967}. The hierarchical cluster has two main components which determines the hierarchy of the clusters, namely the metrics and linkage component. The metrics component is a formula employed for the determination of the distance between points, whilst the linkage is the formula used connect clusters to form a larger clusters. The algorithms chosen for both metric and linkage component are the classic Euclidean \eqref{eq: classic euclidean} \cite{johnson_hierarchical_1967} and complete linkage \eqref{eq: complete linkage} formula \cite{bien2011hierarchical}.

\begin{align}
\label{eq: classic euclidean}
d(\boldsymbol{p},\boldsymbol{q}) = d(\boldsymbol{q}, \boldsymbol{p}) &= \sqrt{(q_1 - p_1)^2 + (q_2 -p_2)^2 + ... + (q_n - p_n)^2}\\
& = \sqrt{\sum_{i}(q_i - p_i)^2}
\end{align}

\begin{equation}
\label{eq: complete linkage}
d(P,Q) = max\{d(\boldsymbol{p},\boldsymbol{q})\} :p \text{ in }P, q \text{ in } B)
\end{equation}

Clustering is applied to the extracted $0^\circ$ gradient points from the detected fractured contours. The result of the clustered points for Figure \ref{fig: Detected Contour Fractures} is illustrated in Figure \ref{fig: Bone Contours Only}. The selected hierarchy cluster is selected based on the distance between the clusters. The selection is performed manually through the GUI. This allows for the user to have precise control over the highlighted fractured regions. Therefore, the application of the clustering technique precisely locates the fractured region within the X-ray image.

\section{Critical Analysis}
\begin{figure}[ht!]
	\centering
	\includegraphics[scale=0.75]{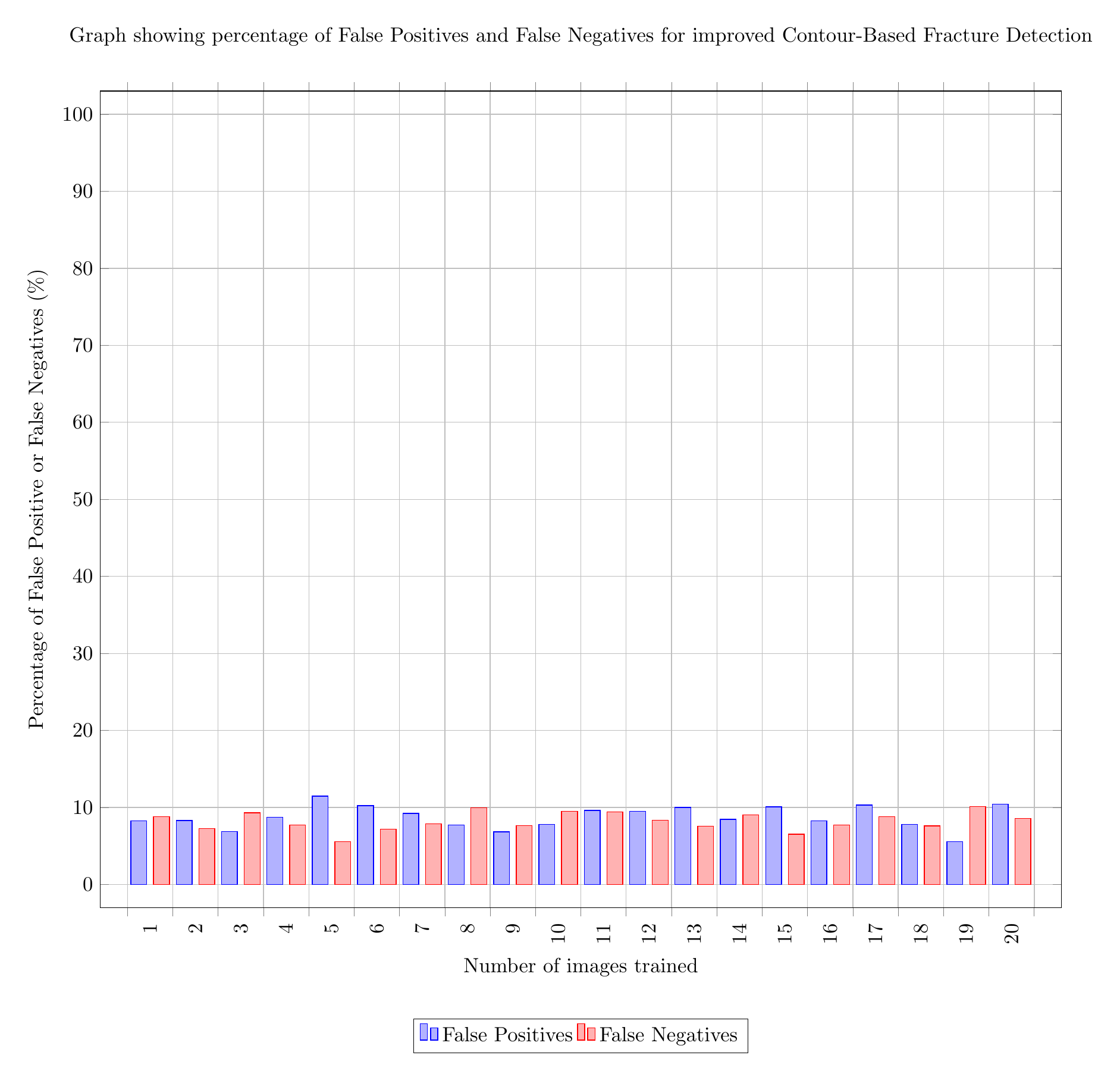}
	\caption{Histogram illustrating the percentage of false positives and false negatives in each case for the improved CHFB fracture detection scheme}
	\label{fig: improved contour false positive and negative ratio}
\end{figure}

The average accuracy for the Standard CHFB scheme is 82.21\%, whilst the improved CHFB scheme 84.03\%. Thus, the improved CHFB scheme performs better than the Standard CHFB scheme. The accuracy for both schemes are calculated using \eqref{eq: accuracy equation}. The detected false positives and negatives from the system are not reflected in the accuracy for both the system and ANN evaluation. The detected false positive and false negatives provides an indication of the systems detection sensitivity. Although, the improved CHFB scheme has a better detection accuracy than the Standard CHFB scheme, it has an equal amount of false positive to false negatives. This is illustrated in Figure \ref{fig: improved contour false positive and negative ratio}. The ratio for false positives to false negatives is 1:0.94, whilst the ratio for Standard CHFB scheme is 1:0.71 for false positive to false negative. However, the improved CHFB scheme has less false detection compared to the Standard CHFB scheme. Majority of the false positive and false negative detections of the CHFB scheme are 10\% and below, whereas the Standard CHFB scheme have false positive detections above 10\%. A detailed presentation of the sensitivity detection is presented in Figures \ref{fig: contour ROC} and \ref{fig: improved contour ROC} for the Standard CHFB and improved CHFB scheme, respectively. The sensitivity and specificity are obtained using \eqref{eq: sensitivity equation} and \eqref{eq: specificity equation}, respectively. From the analysis of Figures \ref{fig: improved contour ROC} and \ref{fig: contour ROC}, the improved CHFB scheme has a better sensitivity with an AUC value of 0.8275, whilst the Standard CHFB scheme has an AUC value of 0.8225. This is despite the flaws of the improved CHFB scheme due to the increased false negatives compared to the Standard CHFB scheme. Therefore, the improved CHFB scheme is opted for compared to the Standard CHFB scheme, as it has an improved accuracy detection and better sensitivity for true positive detection.

\begin{equation}
\label{eq: sensitivity equation}
sensitivity = \frac{\text{TP}}{\text{TP} + \text{FN}}
\end{equation}

\begin{equation}
\label{eq: specificity equation}
specificity = \frac{\text{TN}}{\text{FP} + \text{TN}}
\end{equation}

\begin{equation}
\label{eq: accuracy equation}
a = \frac{\text{TP} + \text{TN}}{\text{TP} + \text{TN} + \text{FP} + \text{FN}}
\end{equation}

\begin{figure}[ht!]
	\centering
	\includegraphics[scale=0.75]{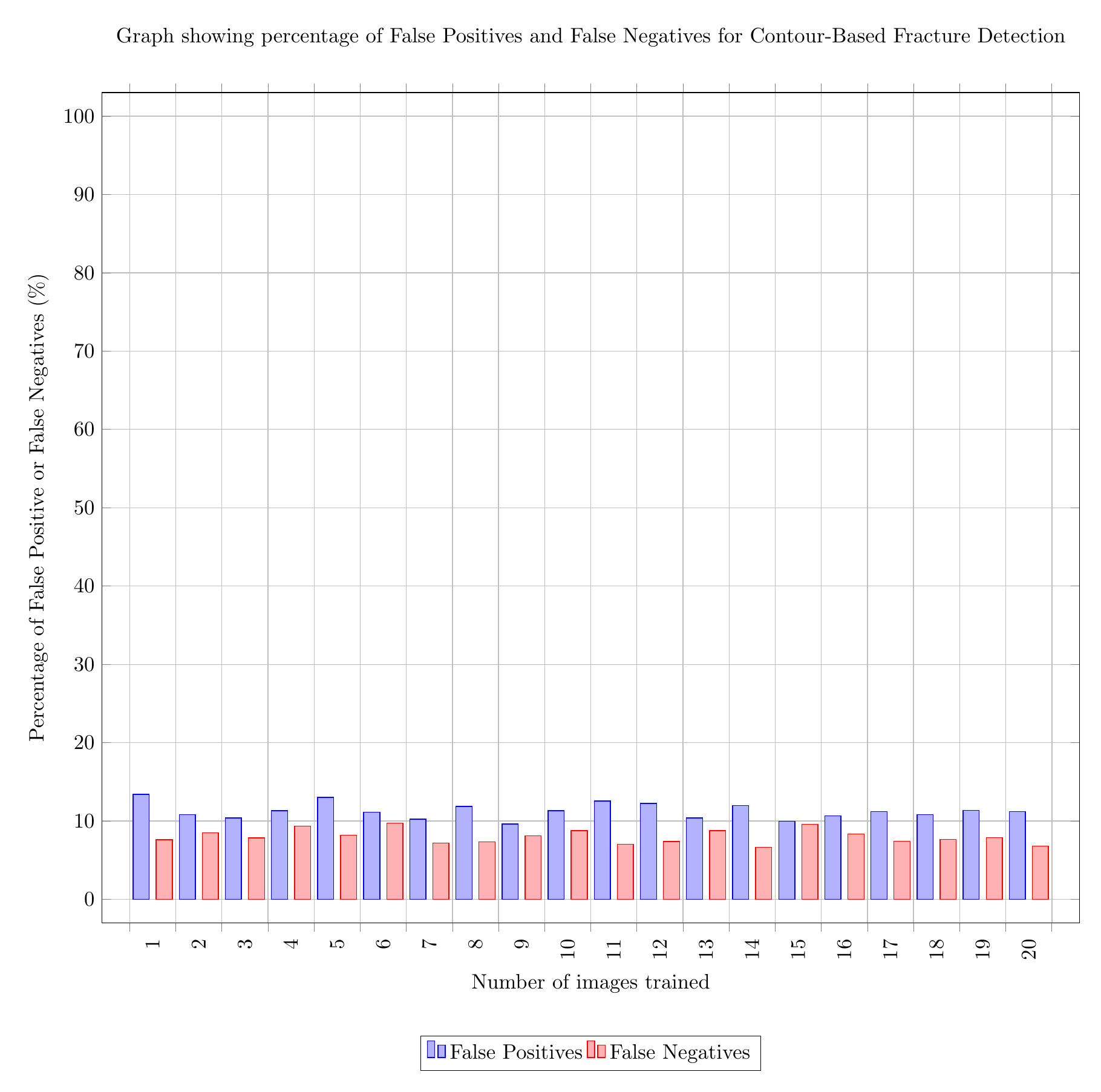}
	\caption{Histogram illustrating the percentage of false positives and false negatives in each case for the Standard CHFB fracture detection scheme}
	\label{fig: contour false positive and negative ratio}
\end{figure}

The accuracy for both systems are hindered by the approach taken to labelling the contours. The area selection approach selects contours based on the contour's presence within the selected area. However contours that are not a fracture, but are within the area are mistakenly labelled as fractured. This occurrence affects a minority of contours, in which it leads to the mislabelled contours utilised for both the training and testing of the ANN. Therefore, with the mislabelling of the minority of the contours it is difficult to obtain a higher accuracy than 89.07\%. An alternative approach for labelling the contour data with more precision is discussed in Section \ref{Section: contour future improvement}.

The refinement of the detection area using the hierarchy clustering approach is limited. As the approach is applied to the extracted $0^\circ$ gradients from detected fractured contours. The limitation is introduced by the Canny edge detection technique, in which the detected edge points are only in four distinct angles, namely $0^\circ$, $45^\circ$, $90^\circ$ and $135^\circ$. The inclusion of any of the other three gradients within the clustering approach introduces an overwhelming amount of details. This causes difficulty in highlighting the fractured region. Therefore, only the $0^\circ$ gradient is useful for the detection of the fractured region. This is problematic as not all fractured contours consist of $0^\circ$ gradients. Thus fractured regions are missed and the user of the system will be required to have some knowledge in referring to the detected fractured contours for fracture detection. Additionally, the clustering method illustrating the precision of the hierarchical cluster level selection can be further improved. This is discussed in Section \ref{Section: contour future improvement}.

\section{Future Improvements} \label{Section: contour future improvement}
\subsection{Data Labelling}
The mislabelling of the contours through the area selection approach, can be improved by introducing a deselection technique within the GUI. This technique allows the user to select individual contours from a small selection of contours that are already considered as fractures. Therefore, this alternative labelling technique consists of both the area selection approach and individual contour selection. Since the individual selection is within a small group of selected contours, it avoids having to select individual contours from all other contour. Hence, this gives the user precise control for contour selection without the task being too cumbersome and time-consuming.

\subsection{Clustering}
The selection of the hierarchical clustering level is provided by the user through the GUI. Although this gives the user full control of the illustrated clustering level for the detection of the fractured region, it requires the user to have some knowledge about the varying clustering levels within a hierarchical cluster. This can be problematic as it would require the user to have prior knowledge about the system. Therefore, an alternative approach is to adopt an adaptive hierarchical clustering level selection technique. The objective of the technique is to optimize the selection of the linkage distances between clusters such that only clusters that highlight the fractured areas within the X-ray image are presented to the user. 

\section{Conclusion}
To conclude, this chapter presents two contour-based fracture detection schemes. The first is a Standard CHFB scheme, in which all extracted contours are labelled and presented to the ANN for the classification of fractured and non-fractured contours. The second is an improved CHFB scheme, whereby further processing is performed to eliminate flesh contours within the leg region from the leg-bone contours. Therefore, the ANN only focuses on contours considered part of the leg-bone. Both schemes are evaluated by its accuracy to accurately classify fractured and non-fractured contours. 

There are a total of 19 features that are extracted from the contours, which are used to provided the ANN with information about individual contours. Two accuracy evaluations are utilised. The first evaluation is a system evaluation, which the ANN is given contours with image context. The second evaluation evaluates the ANN, in which involves contours without any context to its associated image. Both accuracy tests indicate the improved CHFB scheme has a better accuracy ranging from 77.08\% to 87.0\%, whereas the accuracy of the Standard CHFB scheme is between 74.3\% to 85.17\%. However from further analysis of the false positive to false negative ratios for both systems, the Standard CHFB scheme has a ratio of 1:0.71, whereas the improved CHFB scheme has a ratio of 1:0.94. Therefore, the improved CHFB scheme has more false negatives than the Standard CHFB scheme. False negatives within the medical field are not tolerated, however for a system that generalises for varying inputs false negatives are to be minimized. Although the improved CHFB scheme has a higher false negative to false positive ratio, it has a better sensitivity detection compared to the Standard CHFB scheme. This is obtained from analysing the ROC graphs, which depict the sensitivity over the specificity of the system's true positive detection performance. The improved CHFB scheme has an AUC value of 0.8275, whilst the Standard CHFB scheme has a AUC value of 0.8225.

The detected fractured contours are furthered analysed, as the contours do not give a good indication of the fractured region. This is because the fractured contours extend further into the non-fractured regions of the X-ray image. A hierarchical clustering approach is adopted to highlight the fractured regions. This is done by extracting the $0^\circ$ gradients from the detected fractured contours. The maximum accuracy of achieved by the improved CHFB scheme is 89.07\%, however a higher accuracy can be obtained by improving the data labelling approach. The improvement includes an additional methodology in deselecting non-fractured contours that are within the selected fractured area. Furthermore, additional adaptive improvements to the selection of the hierarchical clustering level is considered for future improvements.

\section*{Acknowledgements}
I would like to acknowledge the Radiology department of No. 85 Hospital, Shanghai, China for allowing us to reuse the provided X-ray images for this paper.

\bibliographystyle{IEEEtran}
\bibliography{mybibfile}

\end{document}